\documentclass[11pt]{article}

\usepackage[final]{acl}

\usepackage{times}
\usepackage{latexsym}

\usepackage[T1]{fontenc}
\usepackage[utf8]{inputenc}

\usepackage{microtype}

\usepackage{inconsolata}

\usepackage{graphicx}

\usepackage{amsmath}

\usepackage{booktabs}
\usepackage{multirow}
\usepackage{tabularx}
\usepackage{nicematrix}
\usepackage[table]{xcolor}
\usepackage{float}
\usepackage{tcolorbox}
\usepackage{fvextra}
\usepackage[dvipsnames]{xcolor}
\usepackage{hyperref}

\title{GEM: A Generative Embedding Model Bridging Reasoning and Retrieval}

\author{Zhili Shen \\
  University of Glasgow \\
  \texttt{z.shen.2@research.gla.ac.uk} \\\And
  Craig Macdonald \\
  University of Glasgow \\
  \texttt{craig.macdonald@glasgow.ac.uk}
}

\begin{document}
\maketitle

\begin{abstract}
Modern LLMs excel at reasoning and instruction following, enabling users to express increasingly complex information needs. However, existing retrievers largely rely on matching surface-level relevance, creating a growing gap between the evolving nature of user queries and how the queries are interpreted by retrievers. In this paper, we present GEM, a generative embedding model that augments retrieval by leveraging its internal knowledge to reason about user intent and relevance criteria prior to embedding. GEM unifies generation and embedding in a sequential manner, where encoding can be conceptually viewed as processing only one additional token during autoregressive generation. Unlike conventional query expansion, GEM learns to produce embeddings aligned with its reasoning, moving beyond surface-level matching. Furthermore, the generative nature allows GEM to be prompted like standard instruction-tuned LLMs and enables test-time compute scaling for retrieval. Compared with a non-reasoning variant of the same backbone, GEM achieves approximately $35\%$ relative improvement in nDCG@10 on BRIGHT, a benchmark of challenging queries requiring reasoning, and $70\%$ in p-MRR on FollowIR, a benchmark evaluating retrievers' ability to understand and follow user instructions. Our results demonstrate the effectiveness of activating an embedding model's internal reasoning capabilities to improve retrieval. Our code and model are publicly available.\footnote{
Code: \url{https://github.com/ZhlSh17/GEM}\\
Model: \url{https://huggingface.co/gem-search/GEM-4B}
}

\end{abstract}

\section{Introduction}\label{sec:introduction}
The rapid evolution of large language models (LLMs) is transforming information-access processes. Driven by advanced reasoning \cite{CoT} and instruction-following \cite{FLAN, RLHF} capabilities, these models demonstrate remarkable abilities in understanding diverse inputs and generalising to challenging tasks. Consequently, user behaviour is evolving. For instance, users are increasingly accustomed to interacting with modern AI systems through natural-language instructions \cite{KNOTH2024100225}, which often differ from traditional search queries.

\begin{figure}[t]
    \centering
    \includegraphics[width=\columnwidth]{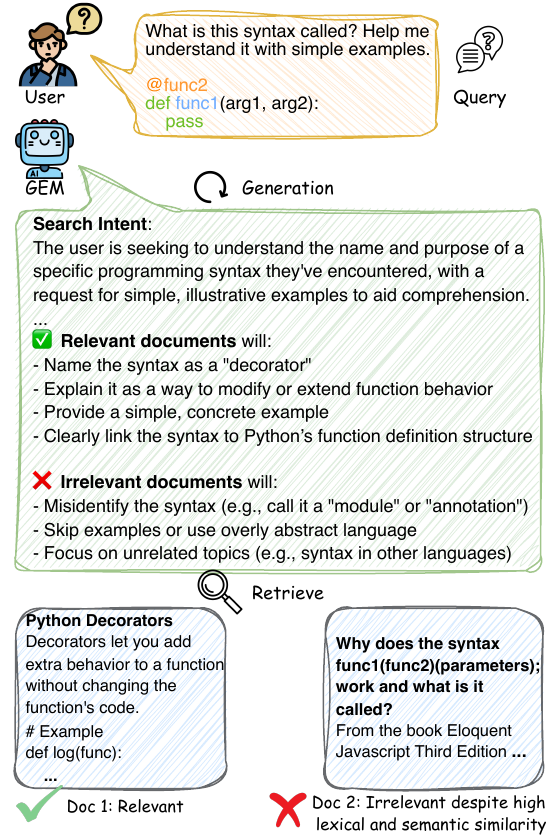}
    \caption{An example use case of GEM.}
    \label{fig:gem_example}
\end{figure}

In contrast, retrievers have not fully kept pace with advances in LLMs. Conventional retrievers primarily rely on lexical and/or semantic similarities between queries and documents \cite{BM25, E5, RepLLaMA, Contriever}. However, addressing the evolving information needs can require deeper query understanding beyond surface-level matching, such as inferring implicit intent \cite{BRIGHT} and interpreting constraints specified in instructions \cite{FollowIR}. This limitation creates a growing gap between how users express their information needs and how retrieval systems interpret them \cite{InfoSearch}.

Recent research has increasingly sought to address these challenges. \citet{BRIGHT} proposed the \textit{reasoning-intensive} retrieval task, where queries are challenging and require reasoning beyond surface-level matching. Follow-up work \cite{ReasonIR, RaDeR, DIVER} has shown that retrievers can be augmented with reasoning generated by upstream LLMs. However, these methods rely on pipelines with separate models for reasoning and retrieval. More importantly, it remains unclear whether the retrievers genuinely understand reasoning, or simply benefit from increased lexical and semantic overlap. In parallel, another line of research has explored \textit{instruction-following} retrieval \cite{FollowIR, InstructIR}. Unlike earlier studies \cite{TART, Instructor} that prepend queries with a task-specific instruction template, recent work enables a dense retriever to represent detailed, query-specific instructions that specify user preferences and constraints \cite{Promptriever}. Yet, user-provided instructions are often underspecified and ambiguous in practice \cite{SteerLLMsViaScalableInteractiveOversight}. Whether reasoning can enhance a retriever's ability to interpret and follow retrieval instructions is underexplored.

In this paper, we present GEM, a \textbf{G}enerative \textbf{E}mbedding \textbf{M}odel. We argue that the challenges outlined above stem from the need to enhance query understanding in retrievers. GEM addresses this by integrating reasoning and embedding within a single model. Figure \ref{fig:gem_example} illustrates GEM's reasoning paradigm: given a query, GEM reasons about the user's intent and relevance criteria. It then encodes the context for retrieval by incrementally processing an additional embedding token. Existing LLM-based embedding models are typically trained as bi-encoders~\cite{Qwen3-Embedding, wang-etal-2024-improving-text, RepLLaMA}; the generative capabilities of their backbones are often degraded due to catastrophic forgetting \cite{CatastrophicForgetting-1, CatastrophicForgetting-2} and/or architectural choices such as bidirectional attention masking \cite{LLM2Vec, LION}. In contrast, GEM is jointly trained with contrastive and causal language modelling objectives, preserving generation while enabling effective embedding. It is thus inherently a \textit{promptable} embedding model that generalises to diverse user instructions. To align embedding with generation, we introduce a tailored data synthesis strategy that constructs document pairs conditioned on validated reasoning: positives align with the reasoning, while hard negatives share similar topics but contain subtle contradictions, thereby discouraging simple surface-level matching.

Our core contributions are: (1) we propose GEM, which unifies representation learning and causal language modelling; (2) we introduce a tailored data generation strategy to train GEM, such that its embedding is aligned with reasoning and therefore interpretable; and (3) our experiments demonstrate the effectiveness of GEM's generative embedding paradigm, with further gains achievable by prompting to exploit test-time compute. To the best of our knowledge, GEM is the first embedding model to leverage its internal generative capabilities to produce reasoning-aligned embeddings for retrieval.

\section{Related Work}\label{sec:related_work}

\paragraph{Reasoning-Intensive Retrieval}
Complex real-world queries can exhibit low lexical and semantic similarity to relevant documents \cite{BRIGHT}. In such cases, effective retrieval requires reasoning beyond surface-level matching. 
% For instance, a patient may describe a set of symptoms, whereas identifying useful medical guidance requires the system to infer a diagnosis.
Prior work, such as ReasonIR \cite{ReasonIR}, trained LLM-based dense retrievers with hard queries. The best performance is often achieved when retrievers are augmented with reasoning generated by upstream LLMs \cite{BRIGHT, ReasonIR, ThinkQE, RaDeR}. However, the alignment between reasoning and retrieval behaviour is unclear: the improvements may stem from additional surface-level overlap rather than genuine understanding of the reasoning (see RQ4 in Section~\ref{sec:results}). GEM is a unified model that leverages its inherent generative capabilities to augment retrieval, with its embedding trained to align with reasoning. Furthermore, GEM's generative nature allows test-time compute scaling through prompting.

\paragraph{Instruction-Following Retrieval}
Training embedding models with instructions aims to improve generalisation across diverse information needs. 
% Prior work~\cite{Instructor, TART} trained embedding models using queries prefixed with fixed, task-specific templates. 
\citet{FollowIR} proposed to evaluate retrievers on queries with flexible, query-specific instructions that can extend beyond topical relevance. For example, a user may prefer documents that are easier to understand and avoid technical jargon. Within this line of research, Promptriever \cite{Promptriever} is a bi-encoder that allows users to specify relevance criteria through prompting (e.g.\ "A document is relevant if ..."). However, relying on user-provided instructions to precisely define the problems is often unrealistic \cite{SteerLLMsViaScalableInteractiveOversight}, for example, due to a lack of domain expertise. Unlike prior work, GEM explicitly reasons about user intent and relevance criteria to augment retrieval. Moreover, GEM is promptable as it preserves the generative capabilities of its backbone and can thus generalise to diverse user instructions.

\paragraph{Query Expansion Using LLMs}\label{sec:related_work_query_expansion}
Query expansion is a well-established technique for improving retrieval performance by mitigating problems such as vocabulary mismatch \cite{GenerativeQRforAdhocSearch}. Prior work such as HyDE \cite{HyDE} and Query2Doc \cite{Query2Doc} leverages the parametric knowledge of LLMs to generate hypothetical documents (answers) for query expansion. While recent studies instead use LLM reasoning~\cite{ThinkQE, RaDeR, O1-Embedder}, they largely follow the conventional query expansion perspective. Our experiments on FollowIR~\cite{FollowIR} reveal a fundamental limitation: retrievers may not understand reasoning over user instructions, such as what not to retrieve (Table~\ref{tab:gem_vs_llm_qe}). In contrast, the role of reasoning in GEM goes beyond bridging vocabulary gaps as in conventional query expansion.

% While conceptually similar to our generate-then-encode paradigm—both involve encoding LLM-generated text for retrieval—there are several key differences. First, our approach unifies generation and embedding within a single model, allowing the embedding computation to reuse the KV cache from generation, thereby improving efficiency. Second, GEM's reasoning may not always correspond to answer-like content. Finally, our model is trained to align its embedding with reasoning. 
\newcommand{\Q}{p}
\newcommand{\R}{r}
\newcommand{\D}{d}

\begin{figure*}[ht]
    \centering
    \includegraphics[width=0.8\textwidth]{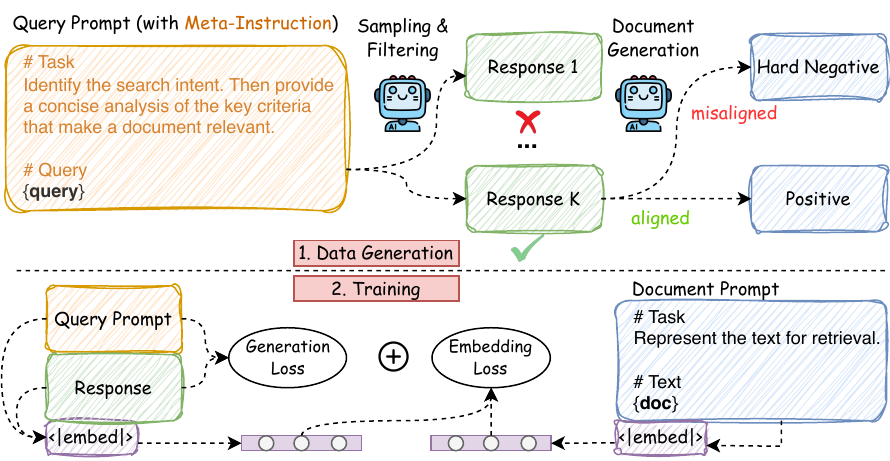}
    \caption{Data generation pipeline (top) and training workflow (bottom) of GEM.}
    \label{fig:gem_method}
\end{figure*}

\section{Generative Embedding Model}
Given a query $q$ and a corpus $\{ d_i \}^N_{i=1}$ containing $N$ documents, the retrieval task aims to find the top-$k$ most relevant documents for $q$, where $k \ll N$. Dense retrieval represents text as embeddings and ranks documents based on the similarity between the query and document embeddings.

Unlike conventional dense retrievers, GEM incorporates generation into its query embedding. Specifically, given a query $q$, we construct a prompt $\Q = I \circ q$, where $I$ is an instruction that prompts GEM to reason about user intent and relevance criteria. We refer to this instruction as our \textit{meta-instruction} to distinguish it from user-provided instructions in queries. GEM then generates a response $r$. For retrieval, GEM learns the similarity between a document $\D$ and the concatenated prompt-response pair, $\Q \circ \R$.

However, incorporating reasoning into embedding raises challenges in both data and training. Regarding data, two key challenges are: (1) the generated reasoning may misinterpret the query, and (2) the need to align the notion of relevance with reasoning. From the training perspective, the model must retain its generative capability while learning effective embeddings. We address these challenges through a tailored data generation pipeline (Section~\ref{sec:method_data}) and joint training with generation and embedding objectives (Section~\ref{sec:method_train}). 

\subsection{Data Generation}\label{sec:method_data}
\paragraph{Response Generation} Retrieval datasets such as MS MARCO \cite{MSMARCO} typically provide a collection of text pairs $\{(q, d^+)\}$ where $d^+$ denotes a relevant (positive) document for query $q$. As illustrated in Figure \ref{fig:gem_method}, given a query $q$, we construct the prompt $\Q$ and sample a set of candidate responses $\mathcal{R}_q = \{ \R^{(1)}, \R^{(2)}, \dots, \R^{(K)} \}$ from an LLM parameterised by $\theta$, i.e.\ $\R^{(k)} \sim P_\theta(\cdot \mid \Q)$. Each response $\R^{(k)}$ may reflect a plausible interpretation of $q$, but could vary in quality and alignment with the original relevance annotations due to hallucinations. To mitigate this, we employ a simple filtering step using an LLM-based relevance classifier. For each candidate response $\R^{(k)}$ and the original positive document $d^+$, we prompt the LLM to assess whether $d^+$ remains relevant under $\R^{(k)}$. Concretely, the classifier outputs a binary decision
$ f(\R^{(k)}, d^+) \in \{0, 1\}$, where $f(\R^{(k)}, d^+) = 1$ indicates that $d^+$ is still relevant given $\R^{(k)}$. We retain only the responses that preserve the relevance of the original positive document:
\begin{equation}
\tilde{\mathcal{R}}_q = \{\R^{(k)} \in \mathcal{R}_q \mid f(\R^{(k)}, d^+) = 1\}.
\end{equation}
If $\tilde{\mathcal{R}}_q$ is empty, the query is discarded from the training set. Otherwise, we randomly select a response $\R \sim \tilde{\mathcal{R}}_q$ to associate with $q$. This filtering step serves as a coarse-grained alignment to avoid obvious contradictions between a generated response and the original positive document.

\paragraph{Document Generation} Original relevance annotations between queries and documents may not fully transfer when conditioning on reasoning. Even after filtering, the relevance of documents may shift subtly due to additional criteria introduced by a response $\R$. To achieve fine-grained alignment, we follow \citet{wang-etal-2024-improving-text} to generate positive and negative documents using an LLM. In our case, the document generation is conditioned on reasoning. For each response $\R$, we prompt the model to generate a positive document based on $\R$. In parallel, hard negative documents may share similar topics but fail to satisfy the user intent or some relevance criteria specified in $\R$. The resulting training set is $\mathcal{T} = \{(\Q, r, d^+, d^-)\}$, where, for simplicity, we reuse $d^+$ and $d^-$ to denote the generated positive and hard negative, respectively. Detailed prompts are provided in Appendix \ref{sec:appendix_prompts}.

\subsection{Training}\label{sec:method_train}
GEM is trained with both causal language modelling and contrastive losses. For each query $q_i$, we construct a prompt $\Q_i$ with the same meta-instruction used for response generation (see Query Prompt in Figure \ref{fig:gem_method}). Given a batch of $N$ prompt-response pairs $\{(\Q_i, \R_i)\}_{i=1}^{N}$, we compute the causal language modelling loss over the responses:
\begin{equation}
    \mathcal{L}_{gen} = - \frac{1}{\sum_{i=1}^N |\R_i|} \sum_{i=1}^N \sum_{t=1}^{|\R_i|} \log P_\theta(\R_{i, t} \mid \Q_i, \R_{i, <t})
\end{equation}
where $\R_{i,t}$ is the $t$-th token of response $\R_i$. 

Embeddings of prompt-response pairs are obtained through last-token pooling, i.e.\ using the hidden state of the last token. Unlike previous studies that use the embedding of the end-of-sequence (EOS) token \cite{RepLLaMA, wang-etal-2024-improving-text}, we reserve the EOS token for generation. Instead, we append a dedicated token, \texttt{<|embed|>}, to the end of each response to produce an embedding. Since generation terminates upon encountering an EOS token, our embedding token is never predicted during decoding and is therefore excluded from the causal language modelling loss. At inference time, the embedding token is appended after generation completes, and its representation is computed efficiently by reusing the KV cache from generation. For document encoding, we prepend each document with a simple instruction (see Document Prompt in Figure \ref{fig:gem_method}) and apply the same embedding token. As generation is not performed on documents, $\mathcal{L}_{gen}$ is not applicable.

Let $E_\theta(\Q_i \circ \R_i)$ be the embedding of the concatenated prompt $\Q_i$ and response $\R_i$, and $E_\theta(\D)$ be the embedding of document $\D$ (with document-side prompt). Following prior work \cite{GritLM, SimCSE, RepLLaMA}, we adopt the InfoNCE loss \cite{InfoNCE} as follows:
\begin{align}
\mathcal{L}_{emb}
= -\frac{1}{N}\sum_{i=1}^N
\log
\frac{
\mathrm{e}^{s_\theta(\Q_i \circ \R_i, \D_i^+)/\tau}
}{
\sum_{\D \in \mathcal{D}_i}
\mathrm{e}^{s_\theta(\Q_i \circ \R_i, \D)/\tau}
}
\end{align}
where $\mathcal{D}_i = \{\D_i^+\} ~ \cup ~ \mathcal{D}_i^- $, and $\mathcal{D}_i^-$ consists of both hard negatives and in-batch negatives. $\tau$ is a temperature hyperparameter. The similarity function $s_\theta$ is defined as:
\begin{align}
s_\theta(\Q_i \circ \R_i, \D)
= \cos\big(E_\theta(\Q_i \circ \R_i), E_\theta(\D)\big)
\end{align}
where $\cos(\cdot, \cdot)$ denotes cosine similarity. Our final training objective is a weighted sum of generation and embedding losses:
\begin{equation}\label{gem_loss}
    \mathcal{L}_{GEM} = \lambda_{gen} \mathcal{L}_{gen} + \lambda_{emb} \mathcal{L}_{emb}
\end{equation}

Overall, our Generative Embedding Model unifies reasoning and embedding within a single model. Our tailored data generation pipeline augments existing retrieval datasets for training GEM and aligns the relevance notion with reasoning. The joint training objective prevents GEM from forgetting token prediction while learning embeddings.
\section{Experiments}
\newcommand{\embedonly}{Qwen3-4B-Instruct}

\subsection{Experimental Setup}\label{sec:experimental_setup}

\paragraph{Evaluation Datasets}
Following~\citet{ReasonIR}, we report nDCG@10 on BRIGHT~\cite{BRIGHT}. BRIGHT comprises 12 datasets spanning diverse domains: Biology (Bio.), Earth Science (Earth.), Economics (Econ.), Psychology (Psy.), Robotics (Rob.), Stack Overflow (Stack.), Sustainable Living (Sus.), LeetCode (Leet.), Pony, mathematics Olympiad problems from AoPS, theorem-based question answering (TheoQ.), and theorem retrieval (TheoT.).

Following~\citet{Promptriever}, we evaluate instruction-following retrieval using FollowIR ~\cite{FollowIR} and InstructIR~\cite{InstructIR} with their official metrics. For FollowIR, we report nDCG@5 on News21, MAP@1000 on Core17 and Robust04, and p-MRR on all splits. As retrievers can ignore instructions while achieving high nDCG or MAP scores on these TREC datasets~\cite{FollowIR}, p-MRR~$\in [-100, 100]$ is proposed to measure the sensitivity to instruction changes, where higher values indicate better instruction following~\cite{Promptriever}. For InstructIR, which is built from MS MARCO~\cite{MSMARCO}, we report nDCG@10 and Robustness@10. The latter is the minimum nDCG@10 for the same query when paired with different prompts, which define distinct user contexts (e.g.\ job)~\cite{InstructIR}.

\paragraph{Implementation Details}
We train GEM from \texttt{Qwen3-4B-Instruct-2507} \cite{Qwen3-Technical-Report}. Experiments with different LLM backbones are presented in Appendix ~\ref{sec:appendix_extra_exp_backbones}. All our models are trained for 500 steps with an effective batch size of 512. The learning rate is set to $1 \times 10^{-5}$ with 50 warmup steps. Following prior work \cite{GritLM, ReasonIR}, we use a training group size of 2, i.e.\ one positive and one hard negative per query. In-batch negatives are then gathered across all GPUs. For training loss, we empirically set $\lambda_{gen} = 0.1$ and $\lambda_{emb} = 1.0$; a parameter study is provided in Appendix~\ref{sec:appendix_param_study}. The temperature $\tau$ for contrastive loss is set to $0.02$. Training takes approximately 14 hours on 2 NVIDIA H100 GPUs.

For data generation, we augment the training data from Promptriever~\cite{Promptriever} and the collection of hard queries from ReasonIR~\cite{ReasonIR} to train GEM. For each query, we sample $K = 8$ responses from GEM's backbone using a temperature of $1.0$. For simplicity, filtering is performed by the same backbone using greedy decoding. For document generation, we use \texttt{Llama-3.1-8B-Instruct} with greedy decoding to generate one positive and one hard negative per response. Additionally, we randomly sample 60,000 instances from the original Promptriever data without applying our data generation pipeline. For these samples, standard contrastive learning is applied, and queries are encoded using the document-side prompt (see Figure~\ref{fig:gem_method}). Our final training set for GEM consists of 370K samples: 320K from Promptriever (comprising 260K with reasoning and 60K original), and 50K reasoning samples based on the hard queries. The size of our 260K reasoning subset is chosen to match our fixed training budget (500 steps with batch size 512), ensuring fair ablation experiments (Table~\ref{tab:ablation}) where models never see repeated samples during training. 

When evaluating GEM, we use greedy decoding throughout all experiments to ensure repeatability and reproducibility. Appendix~\ref{sec:appendix_implementation_details} provides further implementation details.

\begin{table*}[ht]
\centering
\resizebox{\textwidth}{!}{%
\begin{NiceTabular}{l|ccccccc|cc|ccc|c}
\toprule

\multirow{2}{*}{\textbf{Model}} & \multicolumn{7}{c|}{\textbf{StackExchange}} & \multicolumn{2}{c|}{\textbf{Coding}} & \multicolumn{3}{c|}{\textbf{Theorem-based}} & \textbf{Avg.} \\

& \textbf{Bio.} & \textbf{Earth.} & \textbf{Econ.} & \textbf{Psy.} & \textbf{Rob.} & \textbf{Stack.} & \textbf{Sus.} & \textbf{Leet.} & \textbf{Pony} & \textbf{AoPS} & \textbf{TheoQ.} & \textbf{TheoT.} & \\

\midrule

\multicolumn{14}{c}{\cellcolor{gray!10} Single Model} \\
\cline{1-14}

BM25 & 19.2 & 27.1 & 14.9 & 12.5 & 13.5 & 16.5 & 15.2 & 24.4 & 7.9 & 6.0 & 13.0 & 6.9 & 14.8 \\

Contriever & 9.2 & 13.6 & 10.5 & 12.1 & 9.5 & 9.6 & 8.9 & 24.5 & 14.7 & 7.2 & 10.4 & 3.2 & 11.1 \\

% OpenAI & 23.7 & 26.3 & 20.0 & 27.5 & 12.9  & 12.5 & 20.3 & 23.6 & 2.5 &  8.5 & 23.8 & 12.3 & 17.8 \\

GritLM-7B & 25.0 & 32.8 & 19.0 & 19.9 & 17.3 & 11.6 & 18.0 & 29.8 & \textbf{22.0} & 8.8 & 25.1 & 21.1 & 20.9 \\

\embedonly & 18.9 & 35.9 & 18.8 &  24.9 & 18.9 & 20.6 & 17.3 & \textbf{38.5} & 3.8 & 14.4 & 27.1 & 17.8 & 21.4 \\

Promptriever & 25.2 & 38.5 & 21.0  &  25.0 & 19.1  & 18.8 & 18.8  & 32.4 & 1.7 & 9.9  &  21.1 & 8.8  & 20.0  \\

ReasonIR-8B & 26.2 & 31.4 & 23.3 & 30.0 & 18.0 & 23.9 & 20.5 & 35.0 & 10.5 & 14.7 & 31.9 & 27.2 & 24.4 \\

GEM & \textbf{34.7} & \textbf{40.8} & \textbf{26.5} & \textbf{33.7} & \textbf{24.6} & \textbf{29.3} & \textbf{26.0} & 35.5 & 2.8 & \textbf{16.6} & \textbf{37.6} & \textbf{41.8}  & \textbf{29.1} \\

\cline{1-14}
\multicolumn{14}{c}{\cellcolor{gray!10} Pipeline with GPT-4 Reasoner} \\
\cline{1-14}

BM25 & \textbf{53.6} & \textbf{53.6} & 24.3 & 38.6 & 18.8 & 22.7 & 25.9 & 19.3 & 17.7 & 3.9 & 20.2 & 18.9 & 26.5 \\

Contriever & 37.5 & 40.5 & 22.6 & 27.1 & 15.2 & 22.6 & 19.6 & 22.5 & 13.8 & 8.1 & 24.1 & 16.2 & 22.5 \\

GritLM-7B & 33.2 & 33.0 & 23.3 & 30.6 & 15.2 & 17.5 & 21.7 & 33.2 & 11.7 & 6.8 & 26.9 & 28.0 & 23.4 \\

\embedonly &  31.1  & 41.5 & 26.4 & 37.7 & 19.3 & 24.9 & 22.1 & \textbf{36.8} & 2.6 & 10.1 &  33.0  & 31.9  &  26.4 \\

Promptriever & 40.7 &  50.2 & 30.8 & 36.7 & 19.4  & 31.3  &  26.1 & 30.1  & 7.9 & 12.1 & 26.3 & 23.0 & 27.9 \\

Rank1-7B & 48.8 & 36.7 & 20.8 & 35.0 & 22.0 & 18.7 & \textbf{36.2} & 12.7 & 31.2 & 6.3 & 23.7 & 37.8 & 27.5 \\

Rank1-32B & 49.7 & 35.8 & 22.0 & 37.5 & 22.5 & 21.7 & 35.0 & 18.8 & \textbf{32.5} & 10.8 & 22.9 & \textbf{43.7} & 29.4 \\

ReasonIR-8B & 43.6 & 42.9 & \textbf{32.7} & \textbf{38.8} & 20.9 & 25.8 & 27.5 & 31.5 & 19.6 & 7.4 & 33.1 & 35.7 & \textbf{29.9} \\

GEM  & 39.9  & 44.2 & 28.4 & 36.4 & \textbf{25.2} & \textbf{31.2} & 28.1 & 34.7 & 5.2 & \textbf{15.3} & \textbf{36.2} & 35.3 & \textbf{30.0} \\

\bottomrule
\end{NiceTabular}%
}
\caption{nDCG@10 for reasoning-intensive retrieval on BRIGHT. Best results are shown in \textbf{bold}. \embedonly~is the embedding-only variant of GEM with the same 4B-parameter backbone.}
\label{tab:bright}
\end{table*}

\paragraph{Baselines} 
For BRIGHT, we adopt the baseline results reported in ReasonIR. These include: (1) non-LLM retrievers: \textbf{BM25}~\cite{BM25} and \textbf{Contriever}~\cite{Contriever}; (2) LLM-based retrievers: \textbf{GritLM-7B}~\cite{GritLM} and \textbf{ReasonIR-8B}~\cite{ReasonIR}; (3) LLM-based rerankers: \textbf{Rank1-7B} and \textbf{Rank1-32B}~\cite{Rank1} which perform reasoning to rerank the top-$100$ documents retrieved by BM25 with query expansion using GPT-4.

\looseness -1 For instruction-following retrieval, we compile results reported in Promptriever and FollowIR: (1) non-LLM retrievers: \textbf{BM25}~\cite{BM25}, \textbf{BGE-large}~\cite{BGE}, and \textbf{Instructor-XL}~\cite{Instructor}; (2) LLM-based retrievers: \textbf{RepLLaMA}~\cite{RepLLaMA}, \textbf{E5-Mistral}~\cite{wang-etal-2024-improving-text}, \textbf{GritLM-7B}~\cite{GritLM}, and \textbf{Promptriever}~\cite{Promptriever}; (3) LLM-based rerankers: \textbf{FollowIR-7B}~\cite{FollowIR}, \textbf{Mistral-7B-Instruct}~\cite{Mistral-7B-Instruct}, and \textbf{GritLM-7B} (generation mode)~\cite{GritLM}. 

Additionally, we evaluate ReasonIR-8B and Promptriever (7B) across all benchmarks.\footnote{ReasonIR-8B and Promptriever are the closest external baselines to GEM in terms of training data.} To enable a direct comparison using the same backbone, we train \texttt{Qwen3-4B-Instruct-2507} on the original Promptriever data, keeping the training configuration consistent with GEM. This variant is solely an embedding model trained using contrastive loss. We refer to this model as \embedonly. Further ablation studies on the effect of training data will be presented in Table~\ref{tab:ablation}.

\subsection{Results and Analysis}\label{sec:results}
In this section, we present the results and address our main research questions. Additional results are provided in Appendix \ref{sec:appendix_extra_exp}.

\newcounter{rq}
\DeclareRobustCommand{\RQ}{%
  \stepcounter{rq}%
  RQ\arabic{rq}%
}

\paragraph{\RQ: How effective is GEM at reasoning-intensive retrieval?}
Table \ref{tab:bright} presents results on BRIGHT. Our embedding-only variant, \embedonly, serves as a strong baseline, performing comparably to larger models, including GritLM-7B and Promptriever. Overall, GEM achieves an average nDCG@10 of 29.1, outperforming single-model baselines by leveraging its own knowledge. Specifically, GEM excels at theorem-based tasks, substantially improving their average nDCG@10 over \embedonly~($19.8 \rightarrow 32.0$) using the same backbone. However, we note that GEM underperforms on the Pony programming language task. Similar results are observed for Promptriever and \embedonly, despite Promptriever using a different backbone. We provide a detailed discussion in Appendix~\ref{appendix:pony} to explain this out-of-domain problem.\footnote{To address this problem, future work can incorporate the in-domain data provided by \citet{ReasonEmbed} to train GEM.}

Following previous studies \cite{BRIGHT, ReasonIR}, we also report results where retrievers are augmented with reasoning generated by GPT-4, using the official prompt from BRIGHT.\footnote{This data is provided by BRIGHT and is available at: \url{https://huggingface.co/datasets/xlangai/BRIGHT/viewer/gpt4_reason}} In this setting, the average response length is approximately $37\%$ longer than that generated by GEM. Additionally, to investigate whether GEM can effectively represent outputs from a stronger model, we evaluate GEM as a bi-encoder, directly encoding the input instead of generating its own reasoning. With the same reasoning, GEM achieves an average nDCG@10 of 30.0, which is competitive with ReasonIR-8B~(29.9). This observation shows that GEM generalises to such external reasoning. Notably, state-of-the-art embedding models, such as ReasonIR-8B and GritLM-7B, are typically trained using substantially more training compute. In contrast, GEM is a 4B-parameter model trained on two H100 GPUs for approximately 14 hours, serving as an effective demonstration of its generative embedding paradigm at a low training cost.

To answer our RQ1, GEM substantially outperforms the single-model baselines on reasoning-intensive retrieval and generalises to reasoning generated by an external model.

\begin{figure}[tb]
    \centering
    \includegraphics[width=\columnwidth]{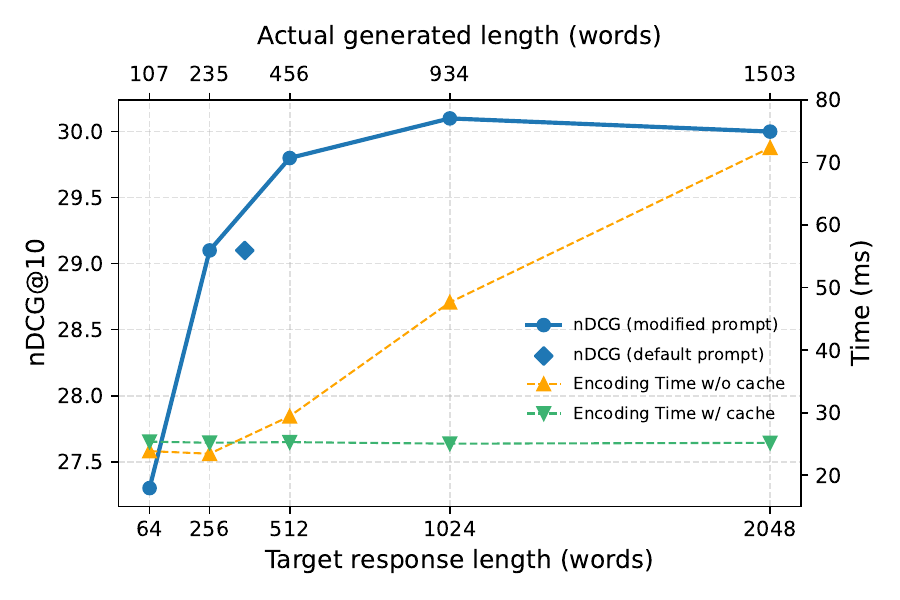}
    \caption{Average nDCG@10 (left) on BRIGHT and per-query encoding time (right) in milliseconds for GEM with test-time compute scaling.}
    \label{fig:gem_ttc_scaling}
\end{figure}

\begin{table*}[ht]
\centering
\resizebox{\textwidth}{!}{%
\begin{tabular}{ll|cc|cc|cc|cc|cc}
\toprule
& \multirow{3}{*}{Model} & \multicolumn{8}{c|}{FollowIR} & \multicolumn{2}{c}{InstructIR}  \\

&  & \multicolumn{2}{c|}{Robust04} & \multicolumn{2}{c|}{News21} & \multicolumn{2}{c|}{Core17} & \multicolumn{2}{c|}{Average} & \multicolumn{2}{c}{MS MARCO}  \\

&  & MAP & p-MRR & nDCG & p-MRR & MAP & p-MRR & Score & p-MRR & nDCG & Robust. \\

 \midrule
\parbox[t]{5mm}{\multirow{3}{*}{\rotatebox[origin=c]{90}{
\parbox[c]{1cm}{\centering \scriptsize Rerankers}}}} & GritLM-7B & 9.7 & +6.1 & 10.2 & +3.4 & 9.8 & +8.6 & 9.9 & +6.0 & -  & - \\

&  Mistral-7B-Instruct & 23.2 & +12.6 & 27.2 & +4.8 & 19.7 & +13.0 & 23.4 & +10.1 & 63.1 & 35.3 \\

& FollowIR-7B & \textbf{24.8} & \textbf{+13.7} & \textbf{29.6} & \textbf{+6.3} & \textbf{20.0} & \textbf{+16.5} & \textbf{24.8} & \textbf{+12.2} & \textbf{81.3} & \textbf{71.5} \\

\midrule
\parbox[t]{5mm}{\multirow{9}{*}{\rotatebox[origin=c]{90}{
\parbox[c]{2cm}{\centering \scriptsize Retrievers}}}} &  BM25 & 12.1 & -3.1 & 19.3 & -2.1 & 8.1 & -1.1 & 13.2 & -2.1 & 76.0 & 26.9 \\

& BGE-large & 17.5 & -7.8 & 22.3 & +0.6 & 15.0 & +0.1 & 18.3 & -2.4 & - & - \\

& Instructor XL & 19.7 & -8.1 & 26.1 & -0.9 & 16.8 & +0.7 & 20.9 & -2.8 & 48.6 & 21.5 \\

\cline{2-12}
& \multicolumn{11}{c}{\cellcolor{gray!10} LLMs $ \geq $7B parameters}\\
\cline{2-12}

& GritLM-7B & \textbf{28.6} & -1.7 & 24.4 & -1.0 & 20.8 & +2.6 & 24.6 & -0.0 & - & - \\

& RepLLaMA & 24.0 & -8.9 & 24.5 & -1.8  & 20.6 & +1.3 & 23.0 & -3.1 & 85.7 & 50.2 \\

& E5-Mistral  &  23.1 & -9.6  & 27.8 & -0.9 & 18.3 & +0.1 & 23.1 & -3.5 & 86.3 &  55.4 \\

& ReasonIR-8B  & 25.3 & -2.8 & 23.1 & +0.6  & 18.7 & +0.9 & 22.4 & -0.4  & 87.1  &  53.1 \\
 
& Promptriever & 28.3 & \textbf{+11.7} & \textbf{28.5} & \textbf{+6.4} & \textbf{21.6} & \textbf{+15.4} & \textbf{26.1} & \textbf{+11.2} & \textbf{92.1} & \textbf{63.1} \\

\cline{2-12}
& \multicolumn{11}{c}{\cellcolor{gray!10} LLMs $\sim$4B parameters}\\
\cline{2-12}

& \embedonly & 25.7 & +7.3 & 25.2 & +3.2 & 21.0 & +9.9 & 24.0 & +6.8 & 82.0 & 46.2 \\

& GEM & \textbf{26.7} & \textbf{+11.6} & \textbf{26.2} & \textbf{+9.3} & \textbf{23.5} & \textbf{+14.2} & \textbf{25.5} & \textbf{+11.7} & \textbf{87.5} & \textbf{54.8} \\

\bottomrule
\end{tabular}%
}
\caption{Results for instruction-following retrieval on the FollowIR and InstructIR datasets. Higher is better for all metrics. MAP@1000, nDCG@5, and Robustness@10 are reported on a 0–100 scale, while p-MRR ranges from -100 to 100. "-" indicates results not reported in previous studies. Best results are shown in \textbf{bold}.}
\label{tab:followir-instructir}
\end{table*}

\paragraph{\RQ: Can GEM’s test-time compute be scaled via prompting, and what is the impact on retrieval performance?}
To investigate whether GEM's retrieval can be influenced by scaling test-time compute, we modify our prompt at test time as follows: (1)~following prior studies \cite{BRIGHT, ReasonIR}, we add an instruction for answer generation, and (2)~we explicitly instruct the model to generate approximately $n \in \{64, 256, 512, 1024, 2048\}$ words. The resulting prompt is provided in Appendix~\ref{sec:appendix_prompts}.

Figure \ref{fig:gem_ttc_scaling} presents the nDCG@10 scores (left) across different target response lengths $n$ (bottom). GEM is aware of our length requirements, adjusting its actual response length~(top) accordingly. Overall, the average nDCG@10 on BRIGHT improves with increased test-time compute, peaking at $30.1$ when prompted with $n = 1024$. However, similar to findings in prior work~\cite{ReasonIR}, performance gains saturate for long generations. We note that this prompting strategy is not flexible, as GEM has to generate redundant content for simple queries to meet our length requirements, while failing to produce additional useful signals for challenging problems beyond its capacity.\footnote{Future work can explore adaptive reasoning with GEM. However, it is beyond the scope of this research question.}

Additionally, Figure \ref{fig:gem_ttc_scaling} benchmarks the average per-query encoding time (right) as $n$ varies. As a baseline, we use GEM to re-encode the same sequence (i.e.\ prompt and response) without reusing the KV cache from generation, analogous to pipelines with independent models. Under this setting, encoding time increases sharply for longer sequences. In contrast, when reusing the KV cache in GEM's generate-then-encode process, encoding time remains stable across the evaluated lengths. Conceptually, GEM's encoding can be viewed as processing one additional token during autoregressive generation, introducing negligible overhead. Compared with systems that rely on separate LLMs, GEM eliminates the need to deploy multiple models and re-encode the same sequence. Appendix~\ref{sec:appendix_latency} provides further latency analysis.

To answer RQ2, GEM supports test-time compute scaling through prompting, with performance on BRIGHT improving as the reasoning budget increases, up to a point of diminishing returns.

\paragraph{\RQ: Can GEM's reasoning augment instruction-following retrieval?}
Table~\ref{tab:followir-instructir} compares GEM with baselines on FollowIR and InstructIR. GEM performs on par with Promptriever and outperforms other LLM-based retrievers on FollowIR. Specifically, GEM achieves a p-MRR of $+11.7$, which indicates strong instruction-following retrieval performance~\cite{Promptriever}, matching the performance of larger models including Promptriever and FollowIR-7B. Compared with the non-reasoning variant \embedonly~(same backbone), GEM achieves notable gains in p-MRR ($+6.8 \rightarrow +11.7$) and Robustness@10 ($46.2 \rightarrow 54.8$). Moreover, disabling reasoning in GEM significantly reduces p-MRR ($+11.7 \rightarrow +9.5$), as shown in Table~\ref{tab:ablation} (i.e.~GEM w/o generation). To answer our RQ3, GEM's reasoning effectively augments its performance on instruction-following retrieval tasks.

As FollowIR is constructed from TREC annotations, its instructions contain detailed relevance criteria (e.g.~"A relevant document will contain \dots"). However, users may not precisely formulate their information needs in practice.
We present qualitative case studies in Appendix~\ref{sec:appendix_examples} to investigate scenarios where such detailed criteria are absent.

\paragraph{\RQ: Can LLM-based query expansion achieve strong instruction-following retrieval like GEM?}
We compare GEM with representative LLM-based query expansion methods, including HyDE~\cite{HyDE} and Query2Doc~\cite{Query2Doc}. For a fair comparison, we use \texttt{Qwen3-4B-Instruct-2507} as the hypothetical document (answer) generator for HyDE and Query2Doc. Implementation details are provided in Appendix \ref{sec:appendix_eval}. Furthermore, we evaluate a variant that concatenates the original query with the reasoning generated by GEM for query expansion.

Results are presented in Table \ref{tab:gem_vs_llm_qe}. While HyDE and Query2Doc often improve nDCG and MAP, their impact on instruction-following metric~(p-MRR) is inconsistent. Notably, both methods degrade p-MRR for Promptriever. While GEM's reasoning contributes to its strong p-MRR, directly applying it to other models may lead to poor results. In answer to our RQ4, LLM-based query expansion is insufficient to achieve strong instruction-following retrieval performance like GEM.

\begin{table}[h]
\centering
\resizebox{\columnwidth}{!}{%
\begin{tabular}{l|cc|cc|cc|ll}

\toprule

\multirow{2}{*}{Setting}  & \multicolumn{2}{c|}{Robust04} & \multicolumn{2}{c|}{News21} & \multicolumn{2}{c|}{Core17} & \multicolumn{2}{c}{Average} \\

& MAP & p-MRR & nDCG & p-MRR & MAP & p-MRR & Score & p-MRR  \\

\midrule

BGE-large & 17.5 & -7.8 & 22.3 & +0.6 & 15.0 & +0.1 & 18.3 & -2.4  \\
\ \ + HyDE  & 21.0 & -4.1 & 23.1 & +0.3 & 20.1 & +1.4 & 21.4$\uparrow$ & -0.8$\uparrow$ \\
\ \  + Query2Doc  & 19.5  & -3.9  & 23.3  & -0.1  & 19.0  & +3.0  & 20.6$\uparrow$ & -0.3$\uparrow$ \\
\ \  + GEM Resp. & 15.8 & -6.2  & 19.9 & -1.9 & 14.8 & -1.0 & 16.9$\downarrow$ & -3.0$\downarrow$ \\

 &  &  &  &  &  &  &  & \\

RepLLaMA & 24.0 & -8.9 & 24.5 & -1.8 & 20.6 & +1.3 & 23.0 & -3.1 \\

\ \ + HyDE & 27.8 & -4.5 & 24.0 & -0.6 & 22.9 & +4.5 & 24.9$\uparrow$ & -0.2$\uparrow$ \\

\ \  + Query2Doc & 27.2 & -2.4 & 22.9 & +0.2 & 21.5 & +2.6 & 23.8$\uparrow$ & +0.1$\uparrow$ \\

\ \  + GEM Resp. & 22.7 & -8.7 & 23.4 & -1.3 & 18.9 & +3.0 & 21.6$\downarrow$ & -2.3$\uparrow$ \\

 &  &  &  &  &  &  &  & \\
 
ReasonIR-8B  & 25.3 & -2.8 & 23.1 & +0.6  & 18.7 & +0.9 & 22.4 & -0.4 \\

\ \ + HyDE  & 26.0 & -1.0 & 22.5 & +1.7 & 21.8 & +4.6 & 23.4$\uparrow$  & +1.7$\uparrow$  \\

\ \ + Query2Doc & 23.1 & -4.4 & 22.8 & +0.4 &  20.6 & +1.0 & 22.2$\downarrow$ & -1.0$\downarrow$ \\

\ \  + GEM Resp. & 19.6  & -8.5 & 22.3 & -0.8 & 18.9 & +0.4 & 20.2$\downarrow$ & -2.9$\downarrow$ \\

 &  &  &  &  &  &  &  & \\

Promptriever & 28.3  & +11.7  & 28.5 & +6.4 & 21.6 & +15.4 & 26.1 & +11.2 \\

\ \ + HyDE & 31.7  & +4.7  & 28.5 & +4.6 & 24.3 & +13.3 & 28.2$\uparrow$ & +7.5$\downarrow$ \\

\ \  + Query2Doc & 30.3 & +6.3  & 26.0  & +3.1 & 23.3 & +13.4 & 26.6$\uparrow$ & +7.6$\downarrow$ \\

\ \  + GEM Resp. & 28.6  &  +10.0 & 30.1 & +3.7 & 22.1 & +15.1 & 26.9$\uparrow$ & +9.6$\downarrow$ \\

 &  &  &  &  &  &  &  & \\

GEM & 26.7 & +11.6 & 26.2 & +9.3 & 23.5 & +14.2 & 25.5 & +11.7 \\

\bottomrule

\end{tabular}%
}
    \caption{Comparisons between GEM and dense retrieval using LLM-based query expansion on the FollowIR dataset. "GEM Resp." means query expansion using GEM's response.  $\uparrow$ indicates improvements over the corresponding baselines; $\downarrow$ denotes declines.}
    \label{tab:gem_vs_llm_qe}
\end{table}

\begin{table}[h]
\centering

\resizebox{\columnwidth}{!}{%
\begin{tabular}{l|l|l}
\toprule
Setting & FollowIR & BRIGHT \\

 & p-MRR & nDCG@10 \\

\midrule

GEM  & +11.7 &  29.1 \\

\cline{1-3}
\multicolumn{3}{c}{\cellcolor{gray!10} Model Ablation} \\
\cline{1-3}

GEM reasoner $+$ GEM encoder (w/o $\mathcal{L}_{gen}$)& +12.5 & 30.0** \\

GEM w/o generation &  +9.5* & 21.0*** \\

\cline{1-3}
\multicolumn{3}{c}{\cellcolor{gray!10} Data Ablation} \\
\cline{1-3}

GEM w/o HQ & +8.5** & 28.4* \\

GEM w/o Orig. Samples & +8.7* & 27.5** \\

GEM w/o HQ and Orig. Samples  & +9.2* & 28.2* \\

GEM w/o Doc. Gen. & +11.7 & 25.8*** \\

\cline{1-3}
\multicolumn{3}{c}{\cellcolor{gray!10} Embedding-Only Variants} \\
\cline{1-3}

\embedonly &  +6.8** & 21.4*** \\

\embedonly \ + HQ  & +5.9*** & 22.0*** \\

\embedonly \ w/ GEM data  & +7.3** & 24.5*** \\

\bottomrule

\end{tabular}%
}
\caption{Ablation study. Following prior work, we report overall scores on FollowIR and BRIGHT. */**/*** denotes a significant difference from GEM at $p < 0.05/0.01/0.001$, respectively, using a two-tailed paired t-test on per-query scores.}
\label{tab:ablation}
\end{table}

\paragraph{\RQ: What is the trade-off of unifying generation and embedding in GEM?} To investigate the trade-off, we train an embedding-only variant of GEM by disabling generation loss on identical training data, referred to as GEM encoder (w/o $\mathcal{L}_{gen}$). Due to catastrophic forgetting, this bi-encoder collapses in next-token prediction, generating repetitive or mixed-language content (see example in Appendix~\ref{sec:appendix_examples}). To evaluate embedding performance under the same encoded sequences, we reuse the reasoning generated by GEM for this embedding-only variant. This setup is analogous to a two-stage pipeline of separate models, and the results are reported in Table~\ref{tab:ablation}. Indeed, our unified model (GEM) shows a slight degradation in p-MRR on FollowIR~($+12.5 \rightarrow +11.7$), and nDCG@10 on BRIGHT~($30.0 \rightarrow 29.1$). However, GEM retains strong generative capabilities and achieves notable improvements on reasoning tasks. Results on generation tasks are provided in Appendix \ref{sec:appendix_extra_exp_generation}. In answer to RQ5, our results show that GEM incurs a modest trade-off in retrieval performance on BRIGHT and FollowIR, while maintaining strong generative capabilities.

\paragraph{\RQ: What are the effects of different components in our training data?} Table~\ref{tab:ablation} presents our ablation study for GEM and \embedonly\ under different training data settings. First, incorporating hard queries (HQ) from ReasonIR into our data generation substantially improves p-MRR on FollowIR ($+8.5 \rightarrow +11.7$), whereas training \embedonly\ with such hard queries (i.e. \embedonly\ + HQ) does not yield similar gains. We postulate that reasoning over these hard queries produces more complex relevance criteria than short factual queries, providing stronger contrastive signals for learning instruction-following retrieval.

Additionally, adding non-reasoning, original samples (Orig.\ Samples) from Promptriever improves both FollowIR p-MRR and BRIGHT nDCG@10. We conjecture that these non-reasoning samples regularise GEM, mitigating overfitting to its reasoning. Interestingly, disabling document generation (i.e.\ GEM w/o Doc. Gen.) substantially reduces nDCG@10 on BRIGHT ($29.1 \rightarrow 25.8$), while FollowIR p-MRR remains stable. This is likely because FollowIR queries contain detailed relevance instructions, which ease alignment given GEM's reasoning. Finally, GEM achieves significant improvements over the embedding-only variant trained on identical samples (i.e.~\embedonly~w/ GEM data).

To answer RQ6, document synthesis substantially improves GEM on BRIGHT, and the inclusion of hard queries and non-reasoning samples yields consistent improvements on both datasets.

\section{Conclusion}
In this paper, we propose GEM, which unifies causal language modelling and representation learning, to bridge the gap between evolving user behaviours and retrievers' ability to understand complex information needs. We present a tailored training data generation method to align GEM's retrieval with reasoning, and a joint training objective to preserve its generative capabilities while learning reasoning-aligned embeddings. GEM achieves strong performance on reasoning-intensive and instruction-following retrieval tasks, despite being a 4B-parameter model trained using substantially less compute than larger baselines. Notably, compared with a controlled non-reasoning variant, GEM achieves approximately 35\% relative improvement in nDCG@10 on BRIGHT and 70\% in p-MRR on FollowIR. We also show that GEM retains its generative capabilities, allowing it to be prompted like standard instruction-tuned LLMs and enabling effective test-time compute scaling for retrieval. Extensive experiments demonstrate the effectiveness of activating GEM's internal reasoning capabilities to enhance retrieval. More broadly, GEM advances the integration of reasoning and embedding, establishing embedding as a native capability in standard LLMs for building more intelligent retrieval systems in future.
\section*{Limitations}
Due to limited computing resources, we are unable to replicate our experiments with larger backbones (e.g.\ 7B parameters) and larger-scale training data. Our model generalisation experiments in Appendix~\ref{sec:appendix_extra_exp_backbones} are restricted to LLM backbones with up to 4B parameters. In contrast to state-of-the-art embedding models, such as GritLM-7B which was trained using $8 \times 8$ A100 GPUs for $3,072$ GPU hours~\cite{GritLM}, this paper provides an effective demonstration of GEM's generative embedding paradigm at a low training cost.

Although our data generation pipeline filters out low-quality responses, hallucinations during document generation and inference are inevitable. Whether hallucinations in generated documents and/or GEM's reasoning can bias its retrieval, and how to quantify the impact, are not investigated.

Similar to existing studies that use LLM reasoning to improve retrieval \cite{ThinkQE, DIVER}, GEM is also limited by the cost of generation. While GEM saves query-side encoding time by reusing the KV cache and allows test-time compute scaling through prompting (see Figure~\ref{fig:gem_ttc_scaling}), autoregressive decoding remains expensive. Further latency discussion is provided in Appendix~\ref{sec:appendix_latency}.

While we note that GEM remains strong on generative tasks (see Appendix~\ref{sec:appendix_extra_exp_generation}), extending it to broader applications that involve retrieval—such as retrieval-augmented generation and conversational search—is valuable for future research. We hope GEM will motivate further exploration of generative embedding models to fill the gaps.

\bibliography{references}

\appendix

\DefineVerbatimEnvironment{VerbatimWrap}{Verbatim}{
breaklines=true, 
breakindent=0pt, 
breaksymbol={},
fontsize=\small,
}

\section{Implementation Details}\label{sec:appendix_implementation_details}

\subsection{Training}\label{sec:appendix_impl_training}
All our models, including those in the ablation study, are trained for 500 steps with the same hyperparameter configuration. We use a per-device batch size of 8 and 32 gradient accumulation steps, which correspond to an effective batch size of 512 on 2 GPUs. Following \cite{GritLM, Instructor}, we use a training group size of 2, i.e.\ every query is paired with one positive and one hard negative. In-batch negatives are then gathered across all GPUs \cite{GritLM}. We use a max length of 1024 tokens for both documents and prompt-response pairs. 

We use the Adam optimiser \cite{Adam}. The learning rate is set to $1 \times 10^{-5}$ with 50 warmup steps. For training loss, we set $\lambda_{gen}$ to $0.1$, and $\lambda_{emb}$ to $1.0$; a parameter study is provided in Appendix~\ref{sec:appendix_param_study}. We empirically used a smaller weight (i.e.\ $\lambda_{gen} = 0.1$) for the generation loss because responses are sampled from the same backbone with the same prompt, and our goal is to preserve the original output distribution while primarily optimising GEM for embedding. The temperature $\tau$ for contrastive loss is set to $0.02$. Although Eq.~\ref{gem_loss} allows decoupled batches for generation and embedding—potentially with different batch sizes—we compute both using a shared batch of prompt-response pairs. We interpret this as a contrastive objective regularised by causal language modelling, which helps prevent the model from forgetting token prediction while learning to encode the same sequences. This design choice also reduces training cost, as the hidden states for computing $\mathcal{L}_{gen}$ can be reused for $\mathcal{L}_{emb}$. Investigating the effects of a decoupled training strategy is left for future work.

To optimise GPU memory usage, we use PyTorch Fully Sharded Data Parallel (FSDP) \cite{FSDP} with CPU offloading and gradient checkpointing. Mixed precision training is enabled with \texttt{bfloat16}. Implementation is supported through Hugging Face Accelerate.\footnote{\url{https://github.com/huggingface/accelerate}}

All our models are trained on a single node with 2 NVIDIA H100 GPUs. For our proposed GEM with \texttt{Qwen3-4B-Instruct-2507} backbone, training completes in approximately 14 hours.

\subsection{Data Generation}\label{sec:appendix_impl_data_generation}

We augment the training data from Promptriever~\cite{Promptriever} and a collection of hard queries from ReasonIR~\cite{ReasonIR} to train GEM. The Promptriever data contains 491K samples from the Tevatron's version\footnote{\url{https://huggingface.co/datasets/Tevatron/msmarco-passage-aug}} of MS MARCO \cite{MSMARCO} and around 500K instruction-rich samples curated by \citet{Promptriever}. These instruction-rich samples were constructed based on MS MARCO queries by adding detailed relevance instructions, and documents were generated using GPT-4. We refer readers to \cite{Promptriever} for details. Note that the non-instructed queries from MS MARCO are typically short and straightforward. The collection from ReasonIR consists of approximately 100K hard queries involving long, challenging problems.

For each query, we sample $K = 8$ responses from GEM's backbone and filter them by validating against the original positives. The prompt for sampling is provided in Figure~\ref{fig:prompt_query}. For filtering, we use the prompt in Figure~\ref{fig:prompt_filter}, where the same LLM backbone is provided with its generated reasoning and the original positive document. The sampling temperature is set to $1.0$. As described in Section~\ref{sec:method_data}, we discard queries with no valid responses. For each of the remaining queries, we randomly select one valid response to associate with them.

For document generation, we employ \texttt{Llama-3.1-8B-Instruct} with greedy decoding to generate one positive and one hard negative per query, given the sampled response. Particularly, the instruction-rich queries from Promptriever contain clearly defined relevance criteria and the corresponding documents (positives and hard negatives) are already generated by GPT-4 based on the criteria. Therefore, we exclude this subset from our document generation and rely on our filtering to ensure the alignment between the responses and GPT-4 generated documents. Figure~\ref{fig:prompt_doc_gen} provides a prompt for document generation. For generating positive documents, we set \texttt{\{relevance\}} to "is relevant". For hard negatives, we randomly assign one of two constraints—"is irrelevant because it fails to satisfy the search intent" or "is irrelevant because it fails to meet some requirements"—each with $50\%$ probability. Following \citet{wang-etal-2024-improving-text}, we generate documents with varying lengths. Instead of specifying an exact word count, we instruct the model with "short", "medium", "long (less than 1000 words)", with probabilities of $50\%$, $45\%$, and $5\%$, respectively. To further improve the diversity of generated documents and mitigate potential hallucinations, we use a modified prompt in Figure~\ref{fig:prompt_doc_gen_with_example} and include one example positive or hard negative document from the original sample. This variant is applied for $50\%$ of the time. Finally, we reuse the filtering prompt from Figure~\ref{fig:prompt_filter} but to filter out generated documents that are inconsistent with our relevance instruction. An example of our document generation is provided in Table \ref{tab:appendix_examples_doc_gen}.

We run both response and document generation on a single NVIDIA H100 GPU using vLLM \cite{PagedAttention}.

Additionally, we randomly sample 60,000 instances from the original Promptriever data. For these non-reasoning samples, standard contrastive learning is applied, and queries are encoded simply using the document-side prompt in Figure \ref{fig:prompt_embed_only}. Our final training set for GEM consists of 370,000 samples: 320,000 from Promptriever (comprising 260,000 with reasoning and 60,000 original), and 50,000 reasoning samples with hard queries from ReasonIR. The size of our 260K reasoning subset is chosen to match our fixed training budget (500 steps with batch size 512), ensuring fair ablation experiments where models never see repeated samples during training. 

For training our embedding-only variant, \embedonly, we sampled 320,000 instances from the original Promptriever data to ensure a consistent ratio with GEM. For our ablation study in Table~\ref{tab:ablation}, we additionally added 50,000 samples (with hard queries) from ReasonIR for training \embedonly, similar to GEM's data setting.

\subsection{Evaluation}\label{sec:appendix_eval}

\paragraph{Setup}
For reproducibility, we use greedy decoding in our evaluation. Unless otherwise specified, we set the maximum generation length to 1024 tokens. The batch size of generate-then-encode is set to 16. For documents and queries when generation is disabled, the encoding batch size is set to 32. For all datasets, both queries and documents are truncated to a maximum length of 1024 tokens. All evaluation experiments are conducted on a single GPU. Unless otherwise specified, we use NVIDIA A6000 for BRIGHT and NVIDIA L40S for FollowIR and InstructIR. To save GPU memory usage, all models are loaded in \texttt{bfloat16}.

\begin{table}[h]
    \centering
    \resizebox{\columnwidth}{!}{%
    \begin{tabular}{c|c|c}
    \toprule
    $n$ & Encoding Time w/o Cache & Encoding Time w/ Cache \\
    \midrule
    64 & 23.87 ± 0.623	& 25.38 ± 0.028   \\
    256 & 	23.45 ± 0.257 & 	25.21 ± 0.021  \\
512 & 	29.44 ± 0.872 & 	25.30 ± 0.035  \\
1024 & 	47.64 ± 0.910 & 	25.06 ± 0.016  \\
2048	 & 72.36 ± 1.333 & 	25.19 ± 0.023 \\
\bottomrule
        
    \end{tabular}%
    }
    \caption{Average per-query encoding time (ms) $\pm$ standard error.}
    \label{tab:encoding_time_with_se}
\end{table}

Particularly, for test-time compute scaling experiments (see Figure \ref{fig:gem_ttc_scaling}), we run models on NVIDIA L40S using a batch size of 1. We randomly sampled 512 queries from BRIGHT to benchmark per-query encoding time under different target length settings. We report the standard errors of encoding time in Table~\ref{tab:encoding_time_with_se}, as they are too small to be visible if plotted in Figure~\ref{fig:gem_ttc_scaling}. The maximum generation length is set to 8192 to ensure sufficient generation budget. For simplicity, the actual generated length reported in Figure~\ref{fig:gem_ttc_scaling} is based on whitespace-separated words rather than tokens.

As described in Section~\ref{sec:experimental_setup}, for our Table~\ref{tab:bright} and~\ref{tab:followir-instructir}, we compile published results from ReasonIR~\cite{ReasonIR}, Promptriever~\cite{Promptriever}, and FollowIR~\cite{FollowIR}.

\paragraph{Evaluation Tools}
For evaluation, we use the official implementations released by the authors of BRIGHT~\cite{BRIGHT}, FollowIR~\cite{FollowIR}, and InstructIR~\cite{InstructIR}, available on GitHub.\footnote{
BRIGHT: \url{https://github.com/xlang-ai/BRIGHT}; \\  FollowIR: \url{https://github.com/orionw/FollowIR}; \\ InstructIR: \url{https://github.com/kaistAI/InstructIR}
} Following their original evaluation protocols, we report results using the standard metrics provided in these codebases. 
% As neither the original implementations nor prior work \cite{Promptriever, ReasonIR} on these benchmarks include statistical significance testing, we do not adapt their evaluation code to incorporate it. Additionally, these codebases build flat indexes for retrieval.

\begin{table*}[t]
\centering

\begin{tabular}{l|ccccc}
\toprule
\multirow{2}{*}{Model} & IFEval  & MMLU & ARC-Challenge  & GSM8K & HumanEval \\

& Acc. & Acc. & Acc. & EM & Pass@1 \\
\midrule
GEM (Qwen3-4B-Instruct) &    84.3 & 65.1 & 48.0 &  82.1 & 89.6 \\

Qwen3-4B-Instruct  & 86.4 & 61.8 & 43.3 &  76.9 & 91.5 \\

Llama3.2-3B-Instruct  & 75.9 & 57.7 & 44.1 & 76.2 & 51.8 \\

Llama3.1-8B-Instruct &  80.7 & 63.1 & 52.7 &  84.4  & 67.1 \\ 

Gemma2-2B-Instruct & 62.5 & 29.1 & 43.5 &  54.1 & 42.1 \\

Qwen2.5-3B-Instruct & 65.9 & 64.6 & 43.0 & 61.3 & 74.4 \\

\bottomrule

\end{tabular}
\caption{Results on generation tasks.}
\label{tab:gen_tasks}
\end{table*}

\section{Additional Experiments}\label{sec:appendix_extra_exp}

\subsection{Results on Generation Tasks}\label{sec:appendix_extra_exp_generation}

To show that GEM maintains its generative capabilities across different tasks, we use Language Model Evaluation Harness\footnote{\url{https://github.com/EleutherAI/lm-evaluation-harness}} (\texttt{lm-eval}) \cite{LM-Eval} to conduct evaluation on the following tasks:

\begin{itemize}
    \item \textbf{IFEval} \cite{IFEval} evaluates a model's instruction-following capability. We report the average accuracy scores across its 4 different settings, including prompt-level strict accuracy, instruction-level strict accuracy, prompt-level loose accuracy, and instruction-level loose accuracy. 
    
    \item \textbf{MMLU} \cite{MMLU} contains knowledge-intensive questions from diverse domains. We report the average accuracy.

    \item \textbf{ARC-Challenge} is the "Challenge" subset from ARC \cite{ARC-Challenge}, consisting of scientific reasoning questions. We report the average accuracy. 
    
    \item \textbf{GSM8K} \cite{GSM8K} focuses on mathematical reasoning. We report the exact match accuracy in the chain-of-thought setting implemented in \texttt{lm-eval}.
    
    \item \textbf{HumanEval} \cite{HumanEval} evaluates Python code generation. We report Pass@1.
    
\end{itemize}

For all these tasks, we use their default configurations from \texttt{lm-eval} except that sampling is disabled to ensure reproducibility.

Table \ref{tab:gen_tasks} presents results from GEM, Qwen3-4B-Instruct (i.e.\ \texttt{Qwen3-4B-Instruct-2507})~\cite{Qwen3-Technical-Report}, Llama3.2-3B-Instruct~\cite{Llama3-Technical-Report}, Llama3.1-8B-Instruct~\cite{Llama3-Technical-Report}, Gemma2-2B-Instruct\footnote{\url{https://huggingface.co/google/gemma-2-2b-it}}~\cite{Gemma2-Technical-Report}, and Qwen2.5-3B-Instruct~\cite{Qwen2.5-Technical-Report}. Overall, while GEM is enabled for embedding, it maintains strong generative performance that is competitive with its backbone, Qwen3-4B-Instruct. In particular, GEM shows substantial gains on the evaluated reasoning benchmarks, improving accuracy on ARC-Challenge from $43.3$ to $48.0$ and EM score on GSM8K from $76.9$ to $82.1$.

\begin{table*}[t]
\centering
\resizebox{\textwidth}{!}{%
\begin{NiceTabular}{l|ccccccc|cc|ccc|c}
\hline
 
\multirow{2}{*}{\textbf{Setting}} & \multicolumn{7}{c|}{\textbf{StackExchange}} & \multicolumn{2}{c|}{\textbf{Coding}} & \multicolumn{3}{c|}{\textbf{Theorem-based}} & \textbf{Avg.} \\

& \textbf{Bio.} & \textbf{Earth.} & \textbf{Econ.} & \textbf{Psy.} & \textbf{Rob.} & \textbf{Stack.} & \textbf{Sus.} & \textbf{Leet.} & \textbf{Pony} & \textbf{AoPS} & \textbf{TheoQ.} & \textbf{TheoT.} & \\ 

\midrule
$\lambda_{gen} = 0.01$ & 31.7 & 36.6 & 27.0 & 31.7 & 23.4 & 29.2 & 25.2 & \textbf{38.3} & \textbf{2.9} & \textbf{17.2} & \textbf{43.6} & 40.0 & 28.9 \\

$\lambda_{gen} = 0.03$ & 32.5 & 38.9 & \textbf{27.7} & 32.0 & 23.9 & 29.1 & 26.4 & 37.2 & 2.4 & 16.6 & 42.7 & 40.7 & 29.2 \\ 

$\lambda_{gen} = 0.1$ (default) & \textbf{34.7} & \textbf{40.8} & 26.5 & \textbf{33.7} & 24.6 & 29.3 & 26.0 & 35.5 & 2.8 & 16.6 & 37.6 & \textbf{41.8}  & 29.1 \\

$\lambda_{gen} = 0.3$ & 33.4 & 39.8 & 27.1 & 32.7 & \textbf{25.1} & \textbf{30.1} & \textbf{29.5} & 36.8 & 1.9 & 16.9 & 37.4 & 40.9 & \textbf{29.3} \\

$\lambda_{gen} = 1.0$ & 30.2 & 35.6 & 25.9 & 30.3 & 23.4 & 26.8 & 27.7 & 35.2 & 1.3 & 15.6 & 36.6 & 38.8 & 27.3 \\

\bottomrule
\end{NiceTabular}%
}
\caption{nDCG@10 on BRIGHT for GEM with different $\lambda_{gen}$. Best results are shown in \textbf{bold}.}
\label{tab:param_study_bright}
\end{table*}

\begin{table*}[t]
\resizebox{\textwidth}{!}{%
\begin{tabular}{l|cc|cc|cc|cc|cc}
\toprule

\multirow{3}{*}{Setting} & \multicolumn{8}{c|}{FollowIR} & \multicolumn{2}{c}{InstructIR} \\

& \multicolumn{2}{c|}{Robust04} & \multicolumn{2}{c|}{News21} & \multicolumn{2}{c|}{Core17} & \multicolumn{2}{c|}{Average} & \multicolumn{2}{c}{MS MARCO} \\

& MAP & p-MRR &  nDCG & p-MRR & MAP & p-MRR & Score & p-MRR & nDCG@10 & Robust.@10 \\
\midrule

$\lambda_{gen} = 0.01$ & \textbf{28.0} &  \textbf{+13.3} &  26.5 &  +8.7 &  \textbf{25.4} & \textbf{ +15.1} & \textbf{26.6} &  \textbf{+12.4} & 83.9 & 49.2 \\

$\lambda_{gen} = 0.03$ & 26.4 & +10.5 & 26.0 & \textbf{+9.9} & 23.7 & +13.7 & 25.4 & +11.4 & \textbf{88.3} & \textbf{56.2} \\

$\lambda_{gen} = 0.1$ (default) &  26.7 & +11.6 & 26.2 & +9.3 & 23.5 & +14.2 & 25.5 & +11.7 & 87.5 & 54.8 \\

$\lambda_{gen} = 0.3$ & 27.0 & +8.7 & 26.9 & +8.3 & 24.9 & +14.4 & 26.3 & +10.5 & 86.8 & 52.2\\

$\lambda_{gen} = 1.0$ & 26.3 & +9.1 & \textbf{27.4} & +7.5 & 23.6 & +14.9 & 25.7 & +10.5 & 87.1 & 53.3 \\

\bottomrule

\end{tabular}%
}
    \caption{Results on FollowIR and InstructIR for GEM with different $\lambda_{gen}$. Best results are shown in \textbf{bold}.}
    \label{tab:param_study_followir_instructir}
\end{table*}

\begin{figure}[h]
    \centering
    \includegraphics[width=\columnwidth]{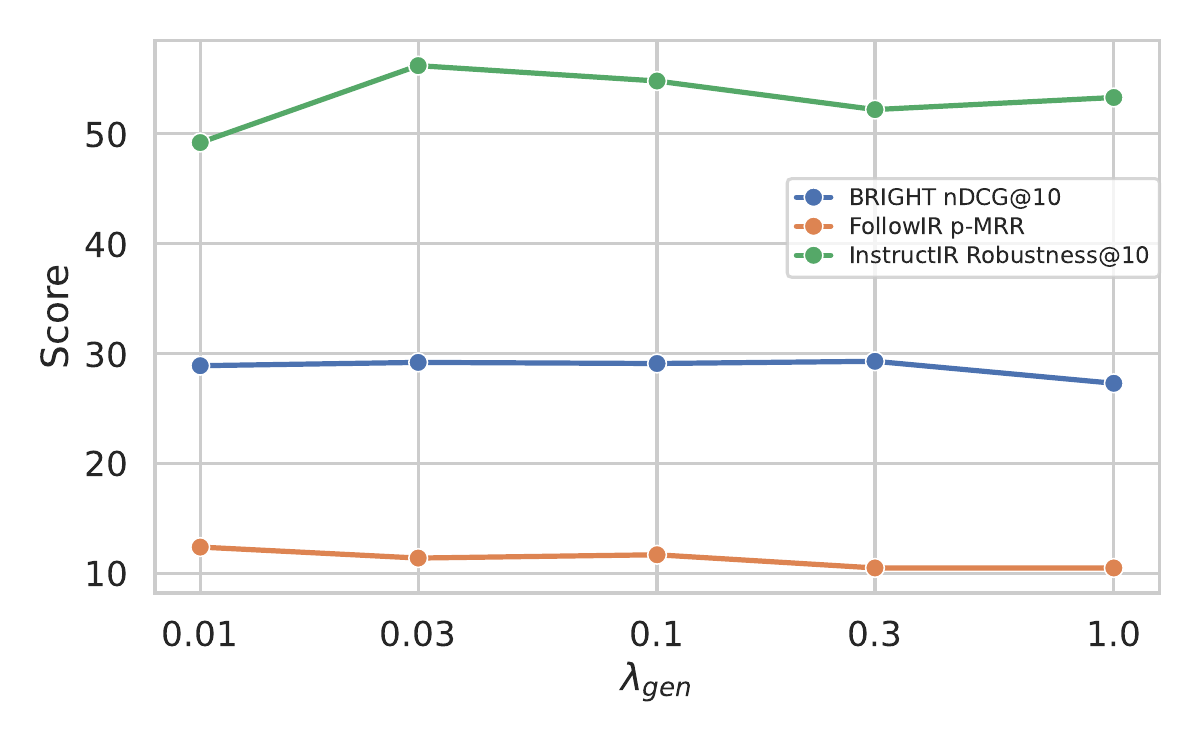}
    \caption{Average nDCG@10 on BRIGHT, p-MRR on FollowIR, and Robustness@10 on InstructIR with $\lambda_{gen} \in \{0.01, 0.03, 0.1, 0.3, 1.0\}$.}
    \label{fig:param_study}
\end{figure}

\subsection{Results with Different Generation Loss Weights}\label{sec:appendix_param_study}

To investigate the effects of $\lambda_{gen}$ for loss weighting (see Eq.~\ref{gem_loss}), we trained GEM with $\lambda_{gen} \in \{0.01, 0.03, 0.1, 0.3, 1.0\}$, where $\lambda_{gen} = 0.1$ is the default setting used in previous experiments.

Table~\ref{tab:param_study_bright} and Table~\ref{tab:param_study_followir_instructir} provide the full results on BRIGHT and instruction-following retrieval benchmarks, respectively. Figure~\ref{fig:param_study} shows the averaged metrics across different values of $\lambda_{gen}$. Overall, we do not observe a clear trend as $\lambda_{gen}$ varies, although the results appear more stable for $\lambda_{gen} \in \{0.03, 0.1, 0.3\}$. These findings suggest that GEM's retrieval performance is relatively insensitive to the exact weighting of the generation loss, provided that $\lambda_{gen}$ lies within a reasonable range.

\begin{table*}[t]
\resizebox{\textwidth}{!}{%
\begin{tabular}{l|cc|cc|cc|cc|cc}
\toprule

\multirow{3}{*}{Model} & \multicolumn{8}{c|}{FollowIR} & \multicolumn{2}{c}{InstructIR} \\

& \multicolumn{2}{c|}{Robust04} & \multicolumn{2}{c|}{News21} & \multicolumn{2}{c|}{Core17} & \multicolumn{2}{c|}{Average} & \multicolumn{2}{c}{MS MARCO} \\

& MAP & p-MRR &  nDCG & p-MRR & MAP & p-MRR & Score & p-MRR & nDCG@10 & Robust.@10 \\
\midrule

\multicolumn{11}{c}{\cellcolor{gray!10} Training with responses from the corresponding backbone} \\
\cline{1-11}

GEM (Qwen3-4B-Instruct) &  26.7 & +11.6 & 26.2 & +9.3 & 23.5 & +14.2 & 25.5 & +11.7 & 87.5 & 54.8 \\

GEM (Llama3.2-3B-Instruct) & 28.9 & +10.9  & 24.5  & +6.5 & 23.1 & +11.4 & 25.5 & +9.6 & 89.8 & 59.0 \\

\cline{1-11}
\multicolumn{11}{c}{\cellcolor{gray!10} Distillation using Qwen3 data} \\
\cline{1-11}

GEM (Llama3.2-3B-Instruct) & 29.3 & +8.3 & 24.8 & +6.4 & 23.7 & +14.1 & 25.9 & +9.6 & 86.4 & 51.2\\

GEM (Gemma2-2B-Instruct) & 22.4 & +13.6 & 25.4 & +11.7 & 21.5 & +11.2  & 23.1 & +12.2 & 84.4 & 48.3 \\

GEM (Qwen2.5-3B-Instruct) & 28.2 & +11.6 & 21.8 & +10.7 & 23.9 & +9.3 & 24.6 & +10.5 & 88.1 & 55.5 \\

\bottomrule

\end{tabular}%
}
    \caption{Results on FollowIR and InstructIR with different backbones.}
    \label{tab:model_generalisation_followir_instructir}
\end{table*}

\begin{table*}[t]
\centering
\resizebox{\textwidth}{!}{%
\begin{NiceTabular}{l|ccccccc|cc|ccc|c}
\hline
 
\multirow{2}{*}{\textbf{Model}} & \multicolumn{7}{c|}{\textbf{StackExchange}} & \multicolumn{2}{c|}{\textbf{Coding}} & \multicolumn{3}{c|}{\textbf{Theorem-based}} & \textbf{Avg.} \\

& \textbf{Bio.} & \textbf{Earth.} & \textbf{Econ.} & \textbf{Psy.} & \textbf{Rob.} & \textbf{Stack.} & \textbf{Sus.} & \textbf{Leet.} & \textbf{Pony} & \textbf{AoPS} & \textbf{TheoQ.} & \textbf{TheoT.} & \\ 

\midrule

\multicolumn{14}{c}{\cellcolor{gray!10} Training with responses from the corresponding backbone} \\
\cline{1-14}

GEM (Qwen3-4B-Instruct) & 34.7 & 40.8 & 26.5 & 33.7 & 24.6 & 29.3 & 26.0 & 35.5 & 2.8 & 16.6 & 37.6 & 41.8  & 29.1 \\

GEM (Llama3.2-3B-Instruct) & 27.1 & 33.9 & 22.3 & 27.4 & 11.9 & 19.5 & 21.6 & 34.0 & 3.1 & 12.9  & 28.3 & 30.5  & 22.7 \\

\cline{1-14}
\multicolumn{14}{c}{\cellcolor{gray!10} Distillation using Qwen3 data} \\
\cline{1-14}

GEM (Llama3.2-3B-Instruct) & 33.2 & 41.7 & 26.7 & 29.4 & 20.5 & 30.4 & 27.1 & 32.6 & 10.0  & 9.9 & 33.8 & 34.6 & 27.5 \\

GEM (Gemma2-2B-Instruct) & 22.4 & 31.2 & 17.5 & 22.9 & 15.6 & 23.3 & 19.7 & 28.1 & 0.49 & 6.4 & 26.9 & 12.3 & 18.9 \\

GEM (Qwen2.5-3B-Instruct) & 28.7 & 35.9 & 20.7 & 29.1 & 17.4 & 25.2 & 23.2 & 35.7 & 2.6 & 11.9 & 35.6 & 27.5 & 24.5 \\

\bottomrule
\end{NiceTabular}%
}
\caption{nDCG@10 on BRIGHT with different backbones.}
\label{tab:model_generalisation_bright}
\end{table*}

\subsection{Results with Different Backbones}\label{sec:appendix_extra_exp_backbones}

We evaluate GEM across different backbones under two settings: (1) the default setting described in Section \ref{sec:method_data}, where reasoning is generated by the corresponding backbone being trained, and (2) a distillation setting in which student models are trained using reasoning generated by \texttt{Qwen3-4B-Instruct-2507}. In the first setting, document generation is also re-run to align with reasoning produced by a different backbone. Particularly, we increase $\lambda_{gen}$ to 0.2 for the distillation setting, as the responses are not sampled from the same backbone. The backbones we used include Qwen3-4B-Instruct~\cite{Qwen3-Technical-Report}, Llama3.2-3B-Instruct~\cite{Llama3-Technical-Report}, Gemma2-2B-Instruct~\cite{Gemma2-Technical-Report}, and Qwen2.5-3B-Instruct~\cite{Qwen2.5-Technical-Report}.

Table \ref{tab:model_generalisation_followir_instructir} provides results on FollowIR and InstructIR. When trained using reasoning generated by the corresponding backbone, GEM (Llama3.2-3B-Instruct) achieves stronger performance on InstructIR. We also observe that models can be effectively trained in the distillation setting, producing results competitive to GEM (Qwen3-4B-Instruct).

Table \ref{tab:model_generalisation_bright} presents results on BRIGHT. Interestingly, for GEM (Llama3.2-3B-Instruct), the distillation setting yields stronger performance than using self-generated reasoning. This suggests that the responses sampled from this backbone may be of comparatively lower quality for BRIGHT. In addition, GEM (Gemma2-2B-Instruct) performs substantially worse on BRIGHT, likely due to its limited reasoning capacity.

\begin{figure}[h]
    \centering
    \includegraphics[width=\columnwidth]{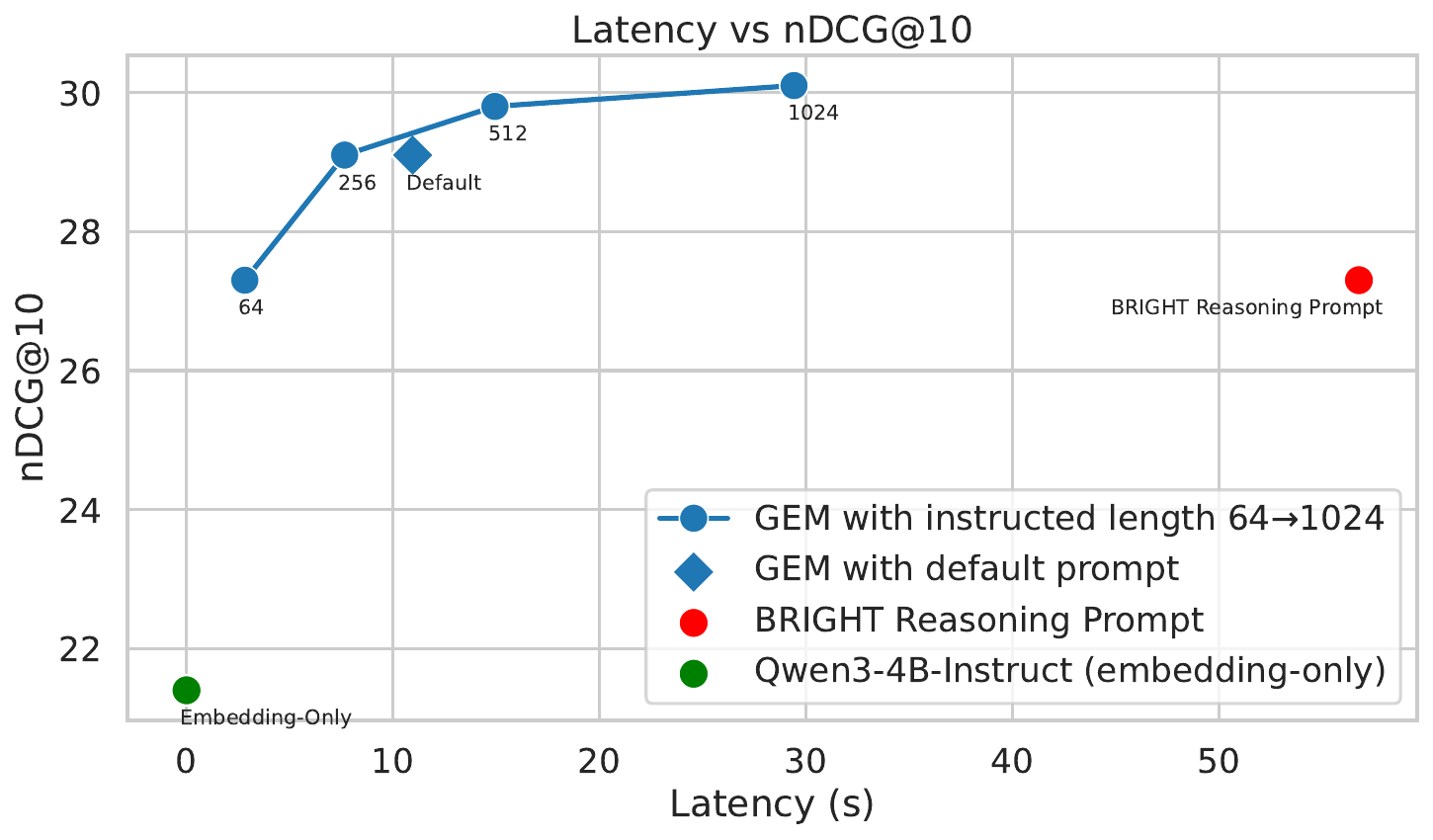}
    \caption{Latency and nDCG@10 on BRIGHT.}
    \label{fig:inference_latency}
\end{figure}

\subsection{Inference Latency}\label{sec:appendix_latency}

To supplement Figure~\ref{fig:gem_ttc_scaling}, we evaluate the combined latency of generation and embedding. Following prior work~\cite{BRIGHT, ReasonIR}, we consider a pipeline comprising an LLM-based reasoner and an LLM-based retriever as our baseline. For reasoning, we adopt the same prompt from BRIGHT, which is available on their official GitHub.\footnote{\url{https://github.com/xlang-ai/BRIGHT/blob/main/reason.py}} This prompt is also adopted in ReasonIR~\cite{ReasonIR}. For a fair comparison with GEM, this baseline uses GEM's backbone, \texttt{Qwen3-4B-Instruct-2507}, as its reasoner. Downstream encoding is performed by our embedding-only variant (i.e.\ \embedonly). All models are benchmarked on the same node with a single NVIDIA L40S GPU. We measure per-query latency by randomly sampling a subset of 128 queries from BRIGHT. We use greedy decoding and a batch size of 1. The maximum length for generation and the truncation length for embedding are increased to 8192.

Figure~\ref{fig:inference_latency} compares the total inference latency and nDCG@10 of different settings on BRIGHT. While reasoning substantially improves nDCG@10 on BRIGHT, autoregressive generation incurs considerable latency overhead compared to the millisecond-scale embedding-only setup (i.e. \embedonly~with original queries). Although GEM allows test-time compute scaling through prompting, the cost of generation remains a limitation, similar to prior studies that rely on LLM-based query expansion to enhance retrieval~\cite{ReasonIR, DIVER, ThinkQE, HyDE, Query2Doc}. Compared with the baseline prompting approach used in BRIGHT and ReasonIR, GEM achieves the same nDCG@10 of 27.3 with a much smaller generation budget ($n=64$), reducing inference latency to 2.85 seconds. This represents approximately a 20$\times$ speedup, demonstrating that GEM's test-time compute scaling can preserve retrieval quality while substantially reducing generation costs.

In practice, GEM's autoregressive generation can be accelerated using well-established optimisation techniques, such as vLLM~\cite{PagedAttention} and FlashAttention~\cite{FlashAttention}. For simplicity and to provide a consistent evaluation setting, we do not incorporate these optimisations. All latency measurements in this paper are evaluated using the default HuggingFace implementation.

\subsection{Evaluation on Additional IR Datasets}\label{appendix:msmarco}

Table~\ref{tab:appendix_msmarco} reports nDCG@10 on TREC-DL-2019~\cite{TREC-DL19} and TREC-DL-2020~\cite{TREC-DL20} with the MS MARCO corpus~\cite{MSMARCO} which contains $8.8$M passages. For GEM, we use the instruction from HyDE~\cite{HyDE}, i.e.~"Please write a passage to answer the question." Compared with the embedding-only variants, GEM improves retrieval performance on these non-reasoning-intensive benchmarks by leveraging its own knowledge. The results are comparable to those of 7B models reported by \citet{Promptriever}.

\begin{table}[!h]
    \centering
    \begin{tabular}{c|c|c}
    \toprule
        Model & DL-2019 & DL-2020 \\
        \multicolumn{3}{c}{\cellcolor{gray!10} 7B Models} \\
        \cline{1-3}
        RepLLaMA & 74.5 & 71.8 \\
        Promptriever & 73.2 & 72.3 \\
        \multicolumn{3}{c}{\cellcolor{gray!10} 4B Models} \\
        \cline{1-3}
        DIVER-4B\footnotemark & 67.8 & 63.6 \\
        \embedonly & 69.0 & 69.1 \\
        GEM w/o generation & 67.4 & 65.8 \\
        GEM & 70.4 & 73.6 \\
        \bottomrule
    \end{tabular}
    \caption{nDCG@10 on DL-2019 and DL-2020.}
    \label{tab:appendix_msmarco}
\end{table}
\footnotetext{\url{https://huggingface.co/AQ-MedAI/Diver-Retriever-4B}}

\subsection{Discussion about Performance on Pony}\label{appendix:pony}
We noted that GEM and Promptriever underperform on the Pony task of BRIGHT (see Table~\ref{tab:bright}). Pony is a programming language, and the corresponding task in BRIGHT has a corpus of approximately 8K documents. We refer readers to the official HuggingFace repository\footnote{\url{https://huggingface.co/datasets/xlangai/BRIGHT/viewer/documents/pony}} for examples. We postulate the problem is because our training queries are largely adopted from Promptriever (MS MARCO), and our data generation pipeline uses LLMs to synthesise training documents. Both the queries and synthesised documents differ substantially in style and content from real ones in the BRIGHT Pony task. Notably, Promptriever itself collapses on Pony ($1.7$ nDCG@10 in Table~\ref{tab:bright}). Furthermore, all GEM variants, regardless of backbone (see Table~\ref{tab:model_generalisation_bright}), perform poorly on Pony, similar to Promptriever. These observations indicate that Pony is out-of-domain with respect to our training data. To address this problem, future work can incorporate the in-domain data provided by \citet{ReasonEmbed} into GEM's training data generation pipeline to improve its performance on BRIGHT.

Additionally, we found that proprietary embeddings reported in the ReasonIR paper \cite{ReasonIR}, including OpenAI, Voyage, and Google,  collapse on Pony as well, with nDCG@10 ranging from $1.5$ to $3.6$.

\section{Prompts}\label{sec:appendix_prompts}

We provide our prompts as follows:
\begin{itemize}
    \item \textbf{Generate-Then-Encode}. Figure~\ref{fig:prompt_query} is our default prompt for generate-then-encode and was used for GEM's data generation and training.

    \item \textbf{Embedding-Only}. Figure~\ref{fig:prompt_embed_only} is the prompt for encoding documents and also queries when generation is disabled.

    \item \textbf{Filtering}. Figure~\ref{fig:prompt_filter} provides our prompt for filtering.

    \item \textbf{Document Generation}. Figure~\ref{fig:prompt_doc_gen} and~\ref{fig:prompt_doc_gen_with_example} provide our document generation prompts without and with an example, respectively. Each prompt is selected with a $50\%$ probability.

    \item \textbf{Test-Time Compute Scaling}. Figure~\ref{fig:prompt_ttc_scaling} provides our prompt for test-time scaling (see Figure \ref{fig:gem_ttc_scaling}). Following BRIGHT~\cite{BRIGHT}, we instruct GEM to generate an answer. Additionally, we explicitly specify the target generation length.

\end{itemize}

\begin{figure}[!h]
\begin{tcolorbox}[
    colback=black!5!white,
    title={Generate-then-encode prompt}
]

\begin{VerbatimWrap}
# Task
Identify the search intent. Then provide a concise analysis of the key criteria that make a document relevant.

# Query
{text}
\end{VerbatimWrap}
\end{tcolorbox}
\caption{Prompt for generate-then-encode.}
\label{fig:prompt_query}
\end{figure}

\begin{figure}[!h]
\begin{tcolorbox}[
    colback=black!5!white,
    title={Embedding-only prompt}
]

\begin{VerbatimWrap}
# Task
Represent the text for retrieval.

# Text
{text}
\end{VerbatimWrap}
\end{tcolorbox}
\caption{Prompt for encoding documents and queries when generation is disabled.}
\label{fig:prompt_embed_only}
\end{figure}

\begin{figure}[!h]
\begin{tcolorbox}[
    colback=black!5!white,
    title={Filtering prompt}
]
\begin{VerbatimWrap}
# 1. User query analysis
{reason}

# 2. Example document
{doc}

# 3. Based on the analysis, is the example document relevant? Just answer Yes or No.
\end{VerbatimWrap}
\end{tcolorbox}
\caption{Prompt for filtering out low-quality data.}
\label{fig:prompt_filter}
\end{figure}

\begin{figure*}[t]
\begin{tcolorbox}[
    colback=black!5!white,
    title={Document generation prompt}
]
\begin{VerbatimWrap}
# Given the following user query analysis
{reason}

# Your task
You are a data annotator. Based on the above analysis, draft a document that {relevance}.

**Important**:
- Preferred length: {length}
- Directly reply with the content **only**
- Do NOT disclose/explain why it is relevant or irrelevant

# Your document
\end{VerbatimWrap}
\end{tcolorbox}
\caption{Prompt for document generation without an example.}
\label{fig:prompt_doc_gen}
\end{figure*}

\begin{figure*}[t]
\begin{tcolorbox}[
    colback=black!5!white,
    title={Document generation prompt with example}
]
\begin{VerbatimWrap}
# Given the following user query analysis
{reason}

# Your task
You are a data annotator. Based on the above analysis, draft a document that {relevance}.

**Important**:
- Preferred length: {length}
- Directly reply with the content **only**
- Do NOT disclose/explain why it is relevant or irrelevant
- You can write in a similar style to the example below; however, make sure your document {relevance}

# Example document
{example}

# Your document
\end{VerbatimWrap}
\end{tcolorbox}
\caption{Prompt for document generation with an example.}
\label{fig:prompt_doc_gen_with_example}
\end{figure*}

\begin{figure*}[t]
\begin{tcolorbox}[
    colback=black!5!white,
    title={Prompt for test-time compute scaling}
]

\begin{VerbatimWrap}
# Task
Identify the search intent. Then think step by step to reason about the criteria that make a document relevant. Finally, draft an answer.
Your response should be approximately {length} words.

# Query
{text}
\end{VerbatimWrap}
\end{tcolorbox}
\caption{Prompt for test-time compute scaling.}
\label{fig:prompt_ttc_scaling}
\end{figure*}

\section{Examples}\label{sec:appendix_examples}

In this section, we provide the following examples:
\begin{itemize}
    \item \textbf{Case Studies.} Table~\ref{tab:appendix_example_doc_ranking}--Table~\ref{tab:appendix_example_doc_ranking_5} present our qualitative case studies, where queries lack detailed relevance criteria. We examine the similarity scores produced by GEM and Promptriever.

    \item \textbf{Response Generation.} Table~\ref{tab:appendix_example_generation_gem_vs_embed_only} illustrates that embedding models suffer from catastrophic forgetting when trained without a causal language modelling objective. In this case, embedding-only models fail to follow our meta-instruction; they generate repetitive content without emitting an EOS token to terminate the generation, although they are trained to use the dedicated token for embedding like GEM.

    \item \textbf{Document Generation.} Table~\ref{tab:appendix_examples_doc_gen} provides an example of our document generation, where the positive document aligns with the reasoning (e.g.\ location and authoritative source), whereas the hard negative fails to meet the search intent despite its topical similarity.
    
\end{itemize}

\newcommand{\green}[1]{\textcolor{ForestGreen}{#1}}
\newcommand{\red}[1]{\textcolor{red}{#1}}

\begin{table*}[t]
    \centering
    \small
    \setlength{\tabcolsep}{4pt}
    
\begin{tabular}{l|p{10cm}}
\toprule

Query  &  

What is this syntax called? Help me understand it with simple examples.

\begin{verbatim}
@func2
def func1(arg1, arg2):
    pass    
\end{verbatim}

\\ \hline

Positive Document & Decorators let you add extra behavior to a function without changing the function's code.

\begin{verbatim}
# Example
def log(func):
   ...    
\end{verbatim}

\\ \hline

Negative Document & 

Why does the syntax func1(func2)(parameters); work and what is it called?
From the book Eloquent Javascript Third Edition ...

\\ \hline

Scores by GEM & Positive Document: \green{0.723} ~ Negative Document: \green{0.570} \\ \hline
Scores by Promptriever & Positive Document: \red{0.578} ~ Negative Document: \red{0.594} \\

\bottomrule
\end{tabular}

\caption{Example of document similarity scores produced by GEM and Promptriever. Scores are highlighted in \green{green} when the positive document is ranked higher and in \red{red} otherwise.}
\label{tab:appendix_example_doc_ranking}
\end{table*}

\begin{table*}[t]
    \centering
    \small
    \setlength{\tabcolsep}{4pt}

\begin{tabular}{l|p{10cm}}
\toprule

Query  &  

"There is always a day when the flower blooms again."

Find me some similar poems.

\\ \hline

Positive Document & 
No winter stays forever.

Somewhere beneath the frost,

a quiet seed is already

dreaming of spring.

\\ \hline

Negative Document & 
Every bloom has its limit. 

The perfume fades to nothing.

Tomorrow brings only empty stems.

The wind carries away the remaining memory.
\\ \hline

Scores by GEM & Positive Document: \green{0.633} ~ Negative Document: \green{0.563} \\ \hline
Scores by Promptriever & Positive Document: \red{0.459} ~ Negative Document: \red{0.471} \\

\bottomrule
\end{tabular}

\caption{Example of document similarity scores produced by GEM and Promptriever. Scores are highlighted in \green{green} when the positive document is ranked higher and in \red{red} otherwise.}
\label{tab:appendix_example_doc_ranking_2}
\end{table*}

\begin{table*}[t]
    \centering
    \small
    \setlength{\tabcolsep}{4pt}
    
\begin{tabular}{l|p{10cm}}
\toprule

Query  &  Explain IBS in simple language. \\ \hline

Positive Document & Irritable Bowel Syndrome (IBS) is a common problem with how your gut works. It can cause stomach pain, bloating, gas, diarrhea, or constipation. IBS does not damage your digestive system, but it makes your gut extra sensitive. Think of it like a miscommunication between your brain and your belly. Certain foods, stress, and lack of sleep can trigger these uncomfortable symptoms. You can often manage IBS by changing your diet, reducing stress, and drinking plenty of water.

\\ \hline

Negative Document & Irritable Bowel Syndrome (IBS) is a chronic gastrointestinal disorder characterized by visceral hypersensitivity and altered intestinal motility. The pathophysiology involves dysregulation of the brain-gut axis, microscopic inflammation, and alterations in the gut microbiota. Quantitative diagnostic criteria, such as the Rome IV parameters, require recurrent abdominal pain associated with defecation or changes in stool frequency and form. Management necessitates targeted pharmacotherapy, including antispasmodics, prokinetics, or secretagogues, to address specific pathophysiological mechanisms.

 \\ \hline

Scores by GEM & Positive Document: \green{0.832} ~ Negative Document: \green{0.676} \\ \hline
Scores by Promptriever & Positive Document: \red{0.430} ~ Negative Document: \red{0.430} \\

\bottomrule
\end{tabular}

\caption{Example of document similarity scores produced by GEM and Promptriever. Scores are highlighted in \green{green} when the positive document is ranked higher and in \red{red} otherwise.}
\label{tab:appendix_example_doc_ranking_3}
\end{table*}

\begin{table*}[t]
    \centering
    \small
    \setlength{\tabcolsep}{4pt}

\begin{tabular}{l|p{10cm}}
\toprule

Query  &  

Which marine organisms do \textbf{not} rely on sunlight for energy?

\\ \hline

Positive Document & Hydrothermal vent communities use chemosynthesis rather than photosynthesis. Bacteria oxidize toxic chemicals like hydrogen sulphide, meaning they do not require any solar energy to survive.

\\ \hline

Negative Document & Phytoplankton and marine algae rely on sunlight for energy to perform photosynthesis. These microscopic organisms form the base of the marine food web by capturing solar radiation.

\\ \hline

Scores by GEM & Positive Document: \green{0.703} ~ Negative Document: \green{0.516} \\ \hline
Scores by Promptriever & Positive Document: \red{0.338} ~ Negative Document: \red{0.391} \\

\bottomrule
\end{tabular}

\caption{Example of document similarity scores produced by GEM and Promptriever. Scores are highlighted in \green{green} when the positive document is ranked higher and in \red{red} otherwise.}
\label{tab:appendix_example_doc_ranking_4}
\end{table*}

\begin{table*}[t]
    \centering
    \small
    \setlength{\tabcolsep}{4pt}
\begin{tabular}{l|p{10cm}}
\toprule

Query  &  

If I have wind speeds at 80m and 120m above the surface and I am interested in approximating the speed at 100m, is it valid to just average the 80 and 120 speeds? In other words, should I expect wind speed to be a linear function of height or would a nonlinear function be more appropriate?

\\ \hline

Positive Document & 
Wind speed increases non-linearly with height due to surface friction and atmospheric stability. Simple linear averaging between 80m and 120m introduces systematic errors because the wind profile follows a logarithmic or power law. To approximate wind speed at 100m accurately, you must use non-linear models like the Logarithmic Wind Profile or the Hellmann power law. These equations account for surface roughness lengths and stability parameters, which cause the velocity curve to shear sharply near the ground and taper off at higher altitudes.

\\ \hline

Negative Document & 
To find the wind speed at 100m using data from 80m and 120m, you can apply a standard linear interpolation formula. Because 100m sits exactly halfway between 80m and 120m, calculating the simple arithmetic mean of the two known velocities provides a highly accurate and valid approximation. Wind velocity increases at a constant, linear rate within this specific boundary layer interval. This straightforward mathematical average eliminates the need for complex, non-linear atmospheric equations or specialized aerodynamic profiling software. \\ \hline

Scores by GEM & Positive Document: \green{0.848} ~ Negative Document: \green{0.680} \\ \hline
Scores by Promptriever & Positive Document: \red{0.707} ~ Negative Document: \red{0.727} \\

\bottomrule
\end{tabular}

\caption{Example of document similarity scores produced by GEM and Promptriever. Scores are highlighted in \green{green} when the positive document is ranked higher and in \red{red} otherwise.}
\label{tab:appendix_example_doc_ranking_5}
\end{table*}

\begin{table*}[t]
    \centering
    \small
    \setlength{\tabcolsep}{4pt}
    \renewcommand{\arraystretch}{0.9}
    
\begin{tabular}{p{3cm}|p{12cm}}
\toprule
Prompt & \# Task \newline
Identify the search intent. Then provide a concise analysis of the key criteria that make a document relevant.
Your response should be around \textbf{100 words}. \newline

\# Query \newline
“There is always a day when the flower blooms again.” \newline
Find me some similar poems.

 \\
\midrule

Response from \\ GEM & **Search Intent:** The user seeks poetic inspiration or similar verses that convey the theme of renewal, resilience, or cyclical beauty—specifically echoing the idea of life returning, like a flower blooming again.  \newline

**Relevance Criteria:** A document is relevant if it features poems with themes of rebirth, persistence, or natural cycles; uses metaphorical language (e.g., flowers, seasons); and expresses hope or continuity. Poems with emotional depth, rhythmic structure, and symbolic imagery align best with the query’s intent. The tone should be reflective, hopeful, or serene to match the original sentiment.
\\ \hline

Response from \\ GEM w/o $\mathcal{L}_{gen}$ & Poem: "The Flower That Blooms Again" by Mary Oliver \newline
Key criteria: cyclical nature, renewal, blooming again, daily recurrence, hope, resilience  
Analysis: This poem captures the recurring cycle of flowers blooming daily, emphasizing hope and resilience through natural repetition. It reflects on the enduring beauty and renewal of life, aligning with the theme of daily renewal and the flower's persistent return. The poem's cyclical imagery and hopefulness match the original quote's essence of continuous blooming and daily hope. cyclic renewal, daily hope, resilience, natural cycles, blooming again. poetic hope, cyclical renewal, daily bloom, resilience, natural cycles. poetic hope, cyclical renewal, daily bloom, resilience, natural cycles. poetic hope, cyclical renewal, daily bloom, resilience, natural cycles \dots
\\ \hline

Response from \\ \embedonly \\ (our embedding-only variant)  & Poems similar to "flower blooms again" include "The Sun Also Rises" and "Daybreak" — both capture renewal and cyclical life. Key criteria: themes of rebirth, daily cycles, and hope. These poems share the poetic structure of nature imagery and recurring daily patterns, mirroring the flower's return. They emphasize resilience and continuity, aligning with the original poem’s intent. Similarity is based on emotional tone, imagery, and cyclical themes. Bloom again poems focus on renewal, not just flowers, but life’s persistence.  

Poems: "Daybreak," "The Sun Also Rises," "Spring Returns" — all reflect daily renewal and hope.  

Criteria: renewal, cyclical nature, hope, daily life.  

Similarity: emotional and thematic resonance with rebirth.  

Flower bloom poems: not just flowers, but life’s return.  

Poems: "Daybreak," "The Sun Also Rises," "Spring Returns" — all reflect daily renewal and hope.  

Criteria: renewal, cyclical nature, hope, daily life.  

Similarity: \dots

\\

\bottomrule
\end{tabular}
\caption{Example responses from GEM and embedding models trained without generation loss. Repetitive content is truncated and denoted by "\dots". We prompt with a response-length requirement to save space. GEM's response contains 123 tokens, while GEM w/o $\mathcal{L}_{gen}$ and \embedonly \ continue generating repetitive content until reaching the maximum generation length in our setting.}
\label{tab:appendix_example_generation_gem_vs_embed_only}
\end{table*}

\begin{table*}[t]
    \centering
    \small
    \setlength{\tabcolsep}{4pt}
    
\begin{tabular}{p{3cm}|p{12cm}}
\toprule
Query & what is the largest lake in oklahoma \\
\midrule

Reasoning & **Search Intent:**  
The user is seeking factual, location-based information about the largest natural lake in the state of Oklahoma. This is a straightforward informational query with a clear geographic and physical attribute focus.

**Relevance Criteria Analysis:**  
A document is relevant if it:  

1. **Accurately identifies the largest lake in Oklahoma** (e.g., Lake Eufaula or Lake Oklahoma, with precise clarification).  

2. Provides **verified geographical and size data** (e.g., surface area in square miles or acres).  

3. Specifies the **location** of the lake within Oklahoma.  

4. Is **authoritative** (e.g., from government sources, reputable environmental organizations, or geographic databases).

5. Avoids confusion with man-made reservoirs or non-lake bodies of water.

*Note:* Lake Thunderbird or Lake Eufaula are often cited, but Lake Eufaula (partially in Oklahoma) is commonly recognized as the largest natural lake in the state by surface area. Accuracy and clarity in distinguishing natural vs. reservoir lakes are critical for relevance.
\\ \hline

Positive & Lake Eufaula is the largest natural lake in Oklahoma, covering an area of approximately 102,000 acres. Located in the southeastern part of the state, it spans across the Oklahoma-Georgia border, with the majority of its surface area situated within Oklahoma. According to the U.S. Army Corps of Engineers, Lake Eufaula has a conservation storage capacity of about 2.3 million acre-feet, making it a vital water resource for the region. \\ \hline

Hard Negative & Discover Lake Pontchartrain. Lake Pontchartrain is a brackish coastal lagoon in Louisiana, covering approximately 630 square miles. It is connected to the Gulf of Mexico and is a popular spot for boating and fishing. The lake is surrounded by the cities of New Orleans and Mandeville. \\

\bottomrule
\end{tabular}
\caption{An example of our generated documents.}
\label{tab:appendix_examples_doc_gen}
\end{table*}

\end{document}